\documentclass[letterpaper]{article} 
\usepackage[preprint]{aaai2027}  
\usepackage[hyphens]{url}  
\usepackage{graphicx} 
\usepackage{natbib}  
\usepackage{caption} 
\usepackage{algorithm}
\usepackage{algorithmic}

\usepackage{amsmath}   
\usepackage{amssymb} 
\usepackage{newfloat}
\usepackage{listings}
\DeclareCaptionStyle{ruled}{labelfont=normalfont,labelsep=colon,strut=off} 
\floatstyle{ruled}
\newfloat{listing}{tb}{lst}{}
\floatname{listing}{Listing}

\usepackage{booktabs}
\usepackage{multirow} 
\usepackage{colortbl}
\usepackage{arydshln}
\definecolor{sealrow}{RGB}{245,245,245}
\definecolor{rl-light}{RGB}{245,245,245}
\title{Guided-RLVR: Guided Exploration in Reinforcement Learning with Verifiable Rewards}
\author{
    Bo Yang\textsuperscript{1,2}\equalcontrib,
    Zishuai Zhang\textsuperscript{1,2}\equalcontrib,
    Weihua Li\textsuperscript{1,2},
    Qinnan Zhang\textsuperscript{1,2},
    Sijia Wen\textsuperscript{1,2},
    Hainan Zhang\textsuperscript{1,2}\corresponding\\
}
\affiliations{
    \textsuperscript{1}Beijing Advanced Innovation Center for Future Blockchain and Privacy Computing, Beihang University\\
    \textsuperscript{2}School of Artificial Intelligence, Beihang University
}

\begin{document}

\newcommand{\MethodName}{Guided-RLVR}
\newcommand{\ControllerName}{Logic Controller}

\maketitle

\begin{abstract}
  Reinforcement learning with verifiable rewards (RLVR) has shown strong empirical gains in improving the reasoning ability of large language models (LLMs) on tasks such as mathematics and programming. However, recent studies have shown that RLVR rollouts often collapse onto a narrow set of reasoning trajectories, limiting the explored reasoning space. Motivated by findings that thinking tokens at critical steps serve as critical structural elements in LLM reasoning pathways, we propose \MethodName{}, a novel approach that augments RLVR with a lightweight controller. The controller strategically injects semantically meaningful steering tokens into the sampling distribution, encouraging the model to approach problems from multiple reasoning perspectives and thereby expanding the exploration space. Experiments on mathematical reasoning and code generation benchmarks show that \MethodName{} consistently outperforms existing RLVR methods while producing more diverse reasoning trajectories.
\end{abstract}

\section{Introduction}

Large language models (LLMs) have demonstrated remarkable capabilities on reasoning-intensive tasks~\cite{achiam2023gpt,team2024gemini}, from mathematical problem solving~\cite{lightman2023let} and code generation~\cite{wei2023magicoder} to multi-step logical inference~\cite{yao2023tree}. While supervised fine-tuning establishes a strong foundation, reinforcement learning with verifiable rewards (RLVR) has emerged as a powerful paradigm for pushing reasoning accuracy further, particularly in domains where correctness verification is available~\cite{shao2024deepseekmath,yu2026dapo}.

Despite these successes, recent work has revealed a fundamental limitation: RLVR predominantly reinforces reasoning paths that already exist in the base model's prior distribution~\cite{tomihari2026power,yue2025does,wu2025invisible}. While this reweighting improves pass@1 accuracy by increasing the probability of correct trajectories, it often fails to expand the \emph{set} of problems the model can solve under repeated sampling. Instead, RLVR-trained models can exhibit \emph{distributional collapse}, in which their rollouts concentrate on a narrow set of high-probability solution patterns and leave large regions of the reasoning space unexplored. This limits the value of inference-time sampling: even with a generous sampling budget, the model may repeatedly generate variations of the same approach rather than discover alternative strategies.

\begin{figure}
  \centering
  \includegraphics[width=\linewidth]{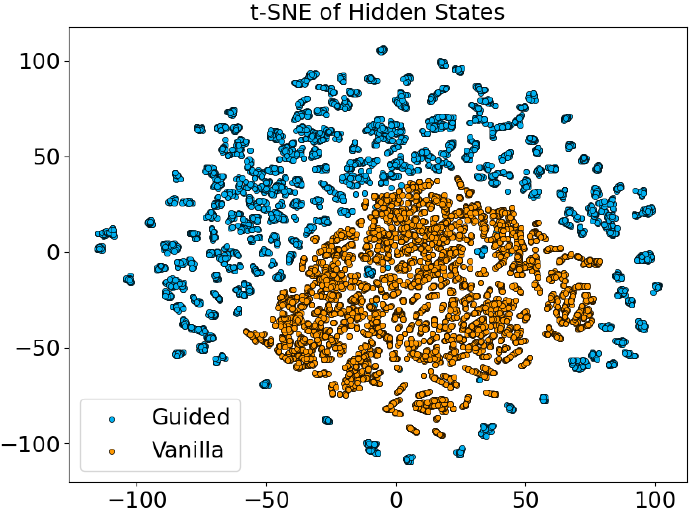}
  \caption{t-SNE~\cite{JMLR:v9:vandermaaten08a} visualization of final-layer rollout representations from vanilla RLVR and guided steering, with 32 rollouts per problem. Total variance is computed in the standardized high-dimensional representation space before projection.}
  \label{fig:motivation}
\end{figure}

We examine this phenomenon by sampling 32 responses per problem on 200 MATH instances~\cite{hendrycks2021measuring} under vanilla RLVR and guided steering. Total variance is computed before dimensionality reduction as the trace of the covariance matrix of the standardized final-layer representations, equivalently the sum of variance across hidden dimensions; t-SNE~\cite{JMLR:v9:vandermaaten08a} is used only for visualization. As shown in Figure~\ref{fig:motivation}, vanilla RLVR produces a compact cluster, whereas guided steering spreads the rollout representations over a substantially broader region and increases total variance by nearly fourfold. This contrast indicates that steering-token interventions reduce the representational concentration observed under vanilla sampling. We observe the same trend with both GRPO~\cite{shao2024deepseekmath} and DAPO~\cite{yu2026dapo} across mathematical reasoning benchmarks.

These observations motivate a more direct approach to expanding the exploration space. In reasoning tasks, certain \emph{thinking tokens}, such as discourse connectives (``Therefore'', ``However'') and reflective markers (``Wait'', ``Since''), carry disproportionately high mutual information with respect to the final answer~\cite{qian2026demystifying}. These tokens appear at structural turning points in reasoning chains, where the model shifts between inductive generalization and deductive inference~\cite{sloman2009causal,cai-etal-2025-role}. Rather than relying solely on stochastic temperature sampling, exploration can therefore be guided by promoting such tokens at selected points during generation.

Building on this insight, we introduce \textbf{\MethodName{}}, which augments RLVR with a lightweight learned \emph{\ControllerName{}}. During rollout generation, the controller is activated at high-entropy sentence boundaries and conditions on the policy's hidden state to select a semantically meaningful steering token. It then boosts the selected token's logit before sampling from the perturbed next-token distribution. The controller and policy are optimized jointly using the same group-relative advantages, reinforcing interventions associated with higher-reward rollouts. Consequently, the guided rollout distribution increases the likelihood of sampling low-prior, high-reward reasoning trajectories that vanilla RLVR rarely explores.

We evaluate \MethodName{} on mathematical reasoning (MATH~\cite{hendrycks2021measuring}, AIME2024, AIME2025, and AMC2023) and code generation (Code-R1) using two backbones (Qwen2.5-7B-Instruct and Qwen3-4B) and two RLVR algorithms (GRPO and DAPO). Across all four combinations of backbone and RLVR algorithm, \MethodName{} achieves the highest average performance, outperforming both vanilla RLVR and the exploration-enhanced methods ENT~\cite{cheng2026reasoning} and REP~\cite{tuyls2026representationbased}. The improvements are particularly consistent at larger sampling budgets. Complementary analyses of trajectory diversity and pass@$k$ scaling show that these gains are accompanied by more diverse reasoning trajectories and broader problem coverage under repeated sampling, supporting the intended role of \MethodName{} in guided exploration.

Our main contributions are:
\begin{itemize}
  \item We show why standard RLVR rarely reaches low-prior trajectories under a finite rollout budget. A union-bound analysis links their vanishing learning signals to their sampling probabilities.

  \item We introduce \MethodName{} with a state-conditioned \ControllerName{}. At high-entropy sentence boundaries, it selects steering tokens to perturb the rollout distribution. The controller and policy are jointly optimized with an auxiliary log-perplexity exploration reward. This design increases the likelihood of sampling low-prior, high-reward reasoning trajectories.

  \item Experiments show that \MethodName{} achieves stronger pass@$k$ scaling and greater trajectory diversity without increasing generation length, suggesting broader exploration of the reasoning space.
\end{itemize}

\section{Related Work}

\subsection{Exploration in RLVR}
RLVR methods such as GRPO~\cite{shao2024deepseekmath} and DAPO~\cite{yu2026dapo} improve reasoning through verifiable feedback, yet gains in single-sample accuracy can coexist with reduced solution coverage under repeated sampling~\cite{yue2025does,wu2025invisible}.
To encourage exploration, ENT~\cite{cheng2026reasoning} augments advantages with a clipped, gradient-detached entropy term, while REP~\cite{tuyls2026representationbased} and DIVER~\cite{hu2026diversityincentivized} introduce representation-based and sequence-level diversity incentives, respectively.
CDE~\cite{dai2025cdecuriositydrivenexplorationefficient} uses response perplexity and critic uncertainty, and likelihood-based rewards provide feedback from reference-answer probabilities~\cite{kwiatkowski2026likelihood}.
These methods shape exploration through optimization signals; our focus is a complementary intervention in the rollout sampling process.

Exploration has also been studied at finer granularity.
\citet{NEURIPS2025_a797c2d2} identify high-entropy forking tokens and concentrate policy-gradient updates on them, alongside decoding experiments at these positions.
SIREN~\cite{jiang2025rethinkingentropy} restricts entropy regularization to selected positions and vocabulary candidates, while Lp-Reg~\cite{huang-etal-2026-low} protects potentially useful low-probability candidates from suppression.
These results motivate targeted exploration rather than indiscriminate entropy maximization.
\MethodName{} builds on this perspective by learning which semantic steering token to promote at high-entropy sentence boundaries, directly biasing the next-token distribution during rollout generation.

\subsection{Guided Rollouts and Reasoning-Time Steering}
G$^2$RPO-A~\cite{guo-etal-2026-g2rpo} and ICPO~\cite{huang-etal-2026-think} guide RL rollouts through reference reasoning steps and in-context expert demonstrations, respectively.
Our controller selects semantic steering tokens from the current hidden state to guide continuations at eligible sentence boundaries.

At inference time, Soft Reasoning~\cite{zhu2025soft} guides generation through first-token embedding search, while ESamp~\cite{zeng2026large} steers decoding using latent novelty estimates.
\citet{qian2026demystifying} use thinking-token interventions to guide reasoning.
\MethodName{} incorporates semantic-token steering into RL rollout generation, jointly optimizing the controller and reasoning policy.
\section{Methodology}

\subsection{Problem Formulation}
Let \(\pi_\theta\) denote the unperturbed LLM policy and \(b_{\mathrm{old}}\) the rollout distribution obtained from its old snapshot using the configured temperature and top-\(k\)/top-\(p\) decoding. Given a prompt \(x\), standard RLVR~\cite{shao2024deepseekmath} samples \(G\) responses
\(\{y^{(i)}\}_{i=1}^{G} \sim b_{\mathrm{old}}(\cdot \mid x)\), computes rewards
\(R_i = R(x, y^{(i)})\), and forms group-relative advantages
\(\hat{A}^{(i)} = (R_i - \mu_R)/(\sigma_R + \epsilon_{adv})\),
where \(\mu_R\) and \(\sigma_R\) are the group mean and standard deviation, and \(\epsilon_{adv}\) is a small constant for numerical stability.
The policy \(\theta\) is updated by minimizing
\begin{equation}
\begin{aligned}
\mathcal{L}_{\text{GRPO}}(\theta)
= -\mathbb{E}_{b_{\mathrm{old}}}\Biggl[
\frac{1}{G}\sum_{i=1}^{G}\frac{1}{T_i}
\sum_{t=1}^{T_i} \\
\min\!\left(r_{i,t}\hat{A}_i,\,
\operatorname{clip}(r_{i,t},1-\epsilon,1+\epsilon)\hat{A}_i\right)
\Biggr],
\end{aligned}
\label{eq:grpo_obj}
\end{equation}
where \(T_i=|y^{(i)}|\), \(r_{i,t}=\pi_\theta(y_t^{(i)}\mid x,y_{<t}^{(i)}) / \pi_{\theta_{\mathrm{old}}}(y_t^{(i)}\mid x,y_{<t}^{(i)})\) is the per-token policy ratio, \(\epsilon\) controls clipping, and no KL penalty is applied. The expectation includes prompts drawn from the training data and rollout groups drawn from \(b_{\mathrm{old}}\). Here \(\theta_{\mathrm{old}}\) denotes the policy snapshot used to collect the rollout group, not the initial pretrained model.

This formulation reweights high-reward trajectories once sampled, but exploration stays tied to the rollout distribution \(b_{\mathrm{old}}\).
Let \(\mathcal{A}_x\) denote valid but low-prior trajectories for \(x\), such as an uncommon decomposition or a corrective reflection step. Each rollout
independently lands in \(\mathcal{A}_x\) with probability \(b_{\mathrm{old}}(\mathcal{A}_x \mid x)\). Let \(P_{\mathrm{hit}}(\mathcal{A}_x)\) denote the probability that at least one of the \(G\) sampled rollouts belongs to \(\mathcal{A}_x\). By Boole's inequality (the union bound),
\begin{align*}
P_{\mathrm{hit}}(\mathcal{A}_x)
&= P\Bigl(\bigcup_{i=1}^{G} \{y^{(i)} \in \mathcal{A}_x\}\Bigr) \\
&\leq \sum_{i=1}^{G} P\bigl(y^{(i)} \in \mathcal{A}_x\bigr)
= G\, b_{\mathrm{old}}(\mathcal{A}_x \mid x).
\end{align*}
When \(b_{\mathrm{old}}(\mathcal{A}_x \mid x) \ll 1/G\), or when it is exactly zero under top-\(p\) truncation, such trajectories are
rarely or never observed under a finite rollout budget. The same bound propagates to the learning signal: since the gradient of
the GRPO objective above is an expectation over \(b_{\mathrm{old}}\), the contribution of \(\mathcal{A}_x\) is scaled by
\(b_{\mathrm{old}}(\mathcal{A}_x \mid x)\) and vanishes in the same edge case (derivation in Appendix~\ref{sec:appendix_gradient_bound}).
Low-prior trajectories thus yield a vanishing learning signal, and unsampled ones none at all: standard RLVR mainly reweights trajectories the
current policy already favors, rather than expanding the explored reasoning space.

This motivates our goal: rather than rely solely on self-sampling from \(b_{\mathrm{old}}\), we construct a guided rollout distribution \(\pi_{\theta_{\mathrm{old}},\phi_{\mathrm{old}}}^{\text{pert}}(\cdot \mid x)\) with the aim of increasing the probability of low-prior, high-reward regions \(\mathcal{A}_x\),
\[
\pi_{\theta_{\mathrm{old}},\phi_{\mathrm{old}}}^{\text{pert}}(\mathcal{A}_x \mid x) > b_{\mathrm{old}}(\mathcal{A}_x \mid x),
\]
while retaining verifiable task feedback. \MethodName{} achieves this with a lightweight controller that perturbs sampling through steering
tokens, guiding exploration beyond the base model's dominant modes.

\subsection{\MethodName{}}
\begin{figure*}[t]
  \centering
  \includegraphics[width=\textwidth]{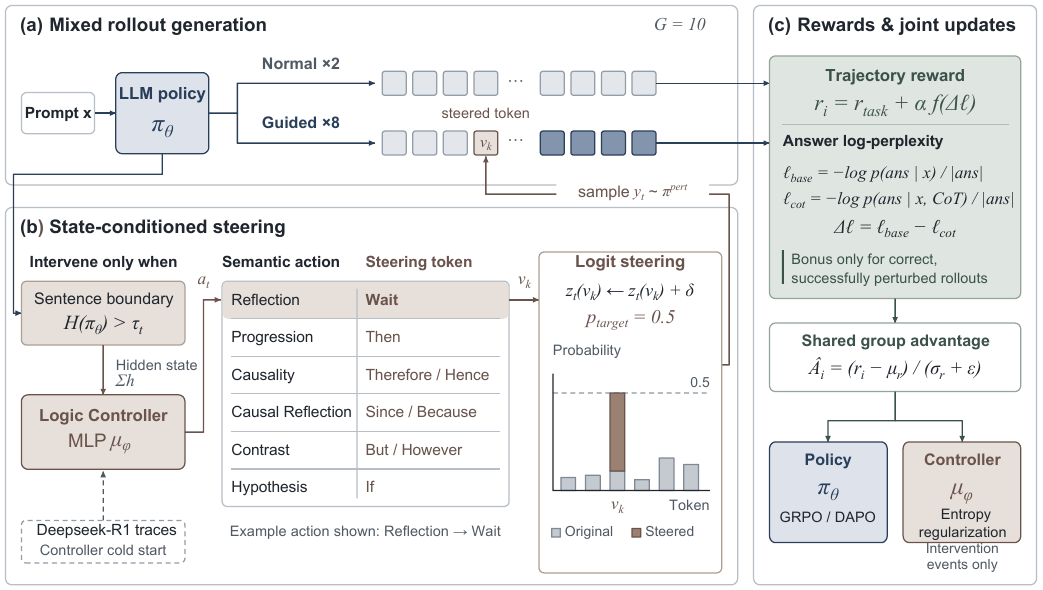}
  \caption{Overview of \MethodName{}.
  \textbf{(a)} For each prompt $x$, policy $\pi_\theta$ generates $G=10$ rollouts: two unperturbed (greedy and stochastic) and eight guided.
  \textbf{(b)} At sentence boundaries where next-token entropy $H(\pi_\theta)$ exceeds the dynamic threshold $\tau_t$, the \ControllerName{} $\mu_\phi$ maps aggregated hidden states $\sum h$ to a semantic action $a_t$ and steering token $v_k$. A bias $\delta$ boosts its logit $z_t(v_k)$ toward probability $p_{\text{target}}=0.5$ before sampling the next token $y_t$ from the perturbed distribution $\pi^{\text{pert}}$.
  \textbf{(c)} Reward $r_i$ combines task correctness $r_{\text{task}}$ with a bonus $\alpha f(\Delta\ell)$, granted only to correct math rollouts where a promoted token was sampled. Here $\Delta\ell=\ell_{\text{base}}-\ell_{\text{cot}}$ measures the same answer's log-perplexity reduction when conditioned on the reasoning chain; $f$ normalizes and clips this difference, and $\alpha$ scales the bonus. Group mean $\mu_r$, standard deviation $\sigma_r$, and stabilizer $\epsilon$ normalize rewards into advantages $\hat{A}_i$ shared by policy and controller updates.}
  \label{fig:framework}
\end{figure*}
Human reasoning combines inductive steps, which infer patterns from instances, and deductive steps, which apply them under constraints~\cite{sloman2009causal}.
This interplay also matters for language-model reasoning: dynamically integrating induction and deduction improves LLM adaptability~\cite{cai-etal-2025-role},
and reflective or transitional tokens (e.g., ``So'', ``Since'', ``If'') coincide with mutual-information peaks that correlate with lower prediction
error~\cite{qian2026demystifying}.

Motivated by these observations, we propose \MethodName{} (Figure~\ref{fig:framework}), a guided RLVR framework with a lightweight controller that perturbs the next-token
distribution toward semantically meaningful steering tokens during rollout. This structured intervention exposes the policy to a broader range of candidate reasoning trajectories.

\subsubsection{\ControllerName{}}
Formally, the \ControllerName{} $\mu_\phi$ is a small MLP that selects which steering token to promote during rollout generation. Its vocabulary and hidden-state aggregation are detailed in Appendix~\ref{sec:appendix_controller_details}.

The controller is queried only when two conditions hold: (i) the previous token ends with one of punctuation marks (\texttt{.} \texttt{?} \texttt{!}), defining a sentence boundary; and (ii) the model exhibits high entropy, indicating uncertainty over the next continuation:
\[
H(\pi_\theta(\cdot \mid x, y_{<t})) > \tau_t,
\]
where $H(\cdot)$ is the entropy and $\tau_t$ is a dynamic threshold (full schedule in Appendix~\ref{sec:appendix_trigger_schedule}).

At a triggered step $t$, the controller conditions on the model's  hidden state and outputs a discrete action
\[
a_t \in \mathcal{A} = \{1,2,\dots,K\},
\]
each corresponding to a steering token $v_k$ from a predefined set $\mathcal{T} = \{v_1, \dots, v_K\}$. The selected token's logit is adjusted by $\delta$ to reach the target probability $p_{\text{target}}$ (derivation in Appendix~\ref{sec:appendix_perturbation}). Temperature scaling and top-$k$/top-$p$ filtering then yield the perturbed distribution $\pi_{\theta,\phi}^{\text{pert}}(y_t \mid x, y_{<t})$.

\subsubsection{Exploration via a Log-Perplexity Reward}
The sparse correctness reward cannot determine whether an intervention contributed to a correct rollout or was merely incidental~\cite{dai2025cdecuriositydrivenexplorationefficient}, so we add a log-perplexity exploration bonus. Following
likelihood-based rewards that use the model's log-probability of a reference answer~\cite{kwiatkowski2026likelihood}, we use the reduction in uncertainty over the answer tokens \(y\) as an auxiliary signal for guided reasoning. Let \(\ell_{\text{base}}\) denote the log-perplexity of the answer sequence \(y\) conditioned only on the prompt \(x\), and let \(\ell_{\text{cot}}\) denote the log-perplexity of the same answer conditioned additionally on the controller-guided reasoning chain \(\hat{x}_{\text{think}}\):
\begin{align*}
\ell_{\text{base}} &= -\frac{1}{|y|}\sum_{i=1}^{|y|} \log p(y_i \mid x, y_{<i}), \\
\ell_{\text{cot}} &= -\frac{1}{|y|}\sum_{i=1}^{|y|} \log p(y_i \mid x, \hat{x}_{\text{think}}, y_{<i}).
\end{align*}
A positive gap \(\Delta \ell = \ell_{\text{base}} -
\ell_{\text{cot}} > 0\) means the reasoning chain reduces the model's uncertainty over the answer. The total reward combines the correctness reward with a scaled version of this exploration bonus:
\[
r_{\text{total}} = r_{\text{orig}} + \alpha \cdot f(\Delta \ell),
\]
where \(f(\cdot)\) normalizes and clips \(\Delta \ell\) (Appendix~\ref{sec:appendix_reward_parameters}).

\subsubsection{Joint Optimization of Policy and Controller}
\label{sec:joint_optimization}

Each rollout group mixes guided completions from $\pi^{\mathrm{pert}}_{\theta_{\mathrm{old}},\phi_{\mathrm{old}}}$ with unperturbed baselines (Section~\ref{sec:setup}). Denote this collection procedure by $\mathcal{Q}_{\mathrm{old}}$ and the group-relative advantage from composite rewards by $\hat{A}_i$.

Logit boosting is used only for rollout sampling. For policy optimization, we keep the sampled completion, including steering tokens, but evaluate both current and old probabilities on the same prefixes using unboosted logits at the configured loss temperature:
\begin{equation}
r_{i,t}(\theta)=
\frac{\pi_\theta(y_t^{(i)}\mid x,y_{<t}^{(i)})}
{\pi_{\theta_{\mathrm{old}}}(y_t^{(i)}\mid x,y_{<t}^{(i)})}.
\label{eq:guided_base_ratio}
\end{equation}
For GRPO, we minimize
\begin{equation}
\begin{aligned}
\mathcal{L}_{\mathrm{LLM}}(\theta)
={}&-\mathbb{E}_{\mathcal{Q}_{\mathrm{old}}}\Biggl[
\frac{1}{G}\sum_{i=1}^{G}\frac{1}{T_i}\sum_{t=1}^{T_i}\\
&\min\!\left(r_{i,t}\hat{A}_i,\,
\operatorname{clip}(r_{i,t})\hat{A}_i\right)\Biggr],
\end{aligned}
\label{eq:guided_llm_objectives}
\end{equation}
where $T_i=|y^{(i)}|$ and clipping uses $[1-\epsilon_{\mathrm{low}},1+\epsilon_{\mathrm{high}}]$. DAPO divides the summed clipped terms by the generation batch's completion-token count. Both objectives use guided data without additional importance correction (Appendix~\ref{sec:policy_update_details}).

The controller shares $\hat{A}_i$, averaging valid successful interventions within each completion and weighting eligible completions equally:
\begin{equation}
\begin{aligned}
\mathcal{L}_{\mathrm{ctrl}}(\phi)
={}&-\mathbb{E}_{i,e}\Bigl[\hat{A}_i\log\mu_\phi(a_e\mid h_e)\\
&+\lambda_H\mathcal{H}\bigl(\mu_\phi(\cdot\mid h_e)\bigr)\Bigr].
\end{aligned}
\label{eq:guided_controller_loss}
\end{equation}
Here $\lambda_H$ weights entropy regularization. The policy and controller use separate optimizers on the same rollout batch.

\section{Experiments}

\subsection{Experimental Setup}
\label{sec:setup}

\subsubsection{Backbone Models}
  We conduct our main experiments on two backbone models: \textbf{Qwen2.5-7B-Instruct}~\cite{qwen2025qwen25technicalreport} and \textbf{Qwen3-4B}~\cite{yang2025qwen3}. These models differ in parameter scale and post-training recipe, allowing us to assess whether the benefits of our method persist across distinct model configurations.

\paragraph{Datasets.} For MATH~\cite{hendrycks2021measuring}, we sample 2,000 training instances and 500 test instances from the original dataset. For Code-R1~\cite{code-r1}, we sample 2,000 training instances and 500 test instances from the original dataset. For the competition benchmarks, we train on 2,000 instances sampled from DAPO-Math-17k-Processed~\cite{yu2026dapo} and evaluate on AIME2024, AIME2025, and AMC2023.

\paragraph{Evaluation metric.} We report pass@$K$~\cite{chen2021evaluating}. Following~\cite{cheng2026reasoning}, we use $K=128$ for the smaller and more challenging competition AIME2024/2025, and AMC2023, and $K=16$ for MATH and Code-R1. The estimator and sampling budgets are detailed in Appendix~\ref{sec:appendix_evaluation}.

\subsubsection{Baselines}
We compare against vanilla GRPO~\cite{shao2024deepseekmath} and DAPO~\cite{yu2026dapo}, as well as two exploration baselines: ENT~\cite{cheng2026reasoning} and REP~\cite{tuyls2026representationbased}. Our method augments GRPO/DAPO with the \ControllerName{}. We hold the RL training data, training schedule, decoding configuration, and evaluation procedure fixed across variants; method coefficients follow the corresponding implementations.

\paragraph{Controller cold start.}
Before RL training, we cold-start the controller on OpenThoughts-114k-math~\cite{guha2025openthoughts} using DeepSeek-R1~\cite{guo2025deepseek} traces, training it to predict six steering categories (Reflection, Progression, Causality, Causal Reflection, Contrast, Hypothesis)~\cite{qian2026demystifying,cai-etal-2025-role} at sentence boundaries (Appendix~\ref{sec:appendix_cold_start}). This provides additional controller supervision that none of the baselines receives. To separate the contribution of supervised initialization from guided RL training, we also evaluate a controller trained from random initialization without this cold start (Section~\ref{sec:ablation}).

Each group contains $G=10$ completions: two unperturbed baselines (one greedy and one stochastic) and eight controller-guided completions. We set $p_{\text{target}}=0.5$ and apply the log-perplexity bonus only to successfully perturbed, correct rollouts. All methods are trained for 5,000 steps on MATH and Code-R1, and 1,000 steps on AIME2024, AIME2025, and AMC2023. Data processing, reward design, and training hyperparameters are detailed in Appendix~\ref{sec:appendix_training}.
\subsection{Main Results}
\label{sec:main_results}

\begin{table*}[t]
\centering
\small
\setlength{\tabcolsep}{4pt}
\renewcommand{\arraystretch}{1.05}
\begin{tabular}{lccccccccccc}
  \toprule
  \multirow{2}{*}{\textbf{Method}} & \multicolumn{2}{c}{\textbf{MATH}} & \multicolumn{2}{c}{\textbf{AIME2024}} & \multicolumn{2}{c}{\textbf{AIME2025}} & \multicolumn{2}{c}{\textbf{AMC2023}} &
\multicolumn{2}{c}{\textbf{Code-R1}} & \multirow{2}{*}{\textbf{AVG}} \\
  \cmidrule(lr){2-3} \cmidrule(lr){4-5} \cmidrule(lr){6-7} \cmidrule(lr){8-9} \cmidrule(lr){10-11}
  & pass@1 & pass@16 & pass@1 & pass@128 & pass@1 & pass@128 & pass@1 & pass@128 & pass@1 & pass@16 & \\
  \midrule
  \rowcolor{sealrow}
  \multicolumn{12}{l}{\textbf{\textit{Qwen2.5-7B-Instruct}}}\\
  GRPO & 62.1 & 79.4 & 11.20 & 46.67 & 5.99 & 50.00 & \textbf{43.23} & 90.36 & 82.0 & 94.8 & 56.58 \\
  GRPO + \textbf{ENT} & 64.5 & 78.8 & 7.97 & 46.67 & \textbf{7.06} & 46.67 & 39.31 & 91.57 & 80.4 & 93.2 & 55.62 \\
  GRPO + \textbf{REP} & 64.9 & 78.0 & 10.89 & 50.00 & 6.48 & 50.00 & 43.22 & 91.57 & \textbf{82.6} & 95.6 & 57.33 \\
  \rowcolor{rl-light} GRPO + \textbf{Ours} & \textbf{65.6} & \textbf{80.6} & \textbf{12.11} & \textbf{53.33} & 5.03 & \textbf{53.33} & 42.77 & \textbf{92.77} & \textbf{82.6} & \textbf{96.0} &
\textbf{58.41} \\
  \hdashline
  DAPO & 65.2 & 81.2 & 10.57& 43.33 & \textbf{5.68} &46.67 & 43.05 & 87.95 & 80.2 & 96.0 & 55.98 \\
  DAPO + \textbf{ENT} & 64.1 & 78.8 & 10.65 & \textbf{50.00} & 4.66  & 46.67 & 42.92 & 90.36 & 81.4 & 94.4 & 56.40 \\
  DAPO + \textbf{REP} & 65.1 & 79.6 & 10.33 & 43.33 & 4.92  & 46.67 & \textbf{43.60} & 89.16 & 81.2 & \textbf{96.8} & 56.07 \\
  \rowcolor{rl-light} DAPO + \textbf{Ours} & \textbf{65.5} & \textbf{81.4} & \textbf{11.56}& \textbf{50.00} & 5.08 & \textbf{50.00} & 42.25 & \textbf{91.57} & \textbf{81.6} & 96.6 & \textbf{57.56} \\
  \midrule
  \rowcolor{sealrow}
  \multicolumn{12}{l}{\textbf{\textit{Qwen3-4B}}}\\
  GRPO  & 65.0 & 78.2 & 15.94 & 63.33 & 12.06 & 53.33 & 43.45 & 93.98 & 85.5 & 95.4 & 60.62 \\
  GRPO + \textbf{ENT} & 63.8 & 77.4 & 25.96 & 73.33 & 20.65 & 70.00 &  56.50& 93.98 & 85.8 & 95.4 & 66.28 \\
  GRPO + \textbf{REP}  & 61.0 & 74.6 & 37.32 & 80.00 & 28.26 &73.33& 66.52 & 96.39 & 85.4 & 95.4 & 69.82 \\
  \rowcolor{rl-light} GRPO + \textbf{Ours}  & \textbf{66.3} & \textbf{78.6} & \textbf{41.17} & \textbf{83.33} & \textbf{32.94} & \textbf{76.67}& \textbf{67.72} & \textbf{97.59} & \textbf{85.9} & \textbf{96.2} & \textbf{72.64} \\
  \hdashline
  DAPO & 66.0 & 76.8 & 14.13 & 73.33 & 13.20 & 70.00 & 45.52 & 92.77 & 85.6 & 94.8 & 63.21 \\
  DAPO + \textbf{ENT} & 64.4 & 75.8 & 20.29 & 76.67 & 19.09 & 70.00 & 56.04 & 95.18 & 85.6 & 96.0 & 65.91 \\
   DAPO + \textbf{REP} & 62.5 & 74.4 &  38.20 & \textbf{83.33} & 29.19 & 73.33 & \textbf{64.05} & 97.59 & 85.2 & \textbf{96.2} & 70.40 \\
   \rowcolor{rl-light} DAPO + \textbf{Ours} & \textbf{66.5} & \textbf{77.4} & \textbf{38.98} & \textbf{83.33} & \textbf{29.84}& \textbf{76.67} & 63.38 & \textbf{98.80} & \textbf{85.8} & 96.0 & \textbf{71.67} \\
  \bottomrule
\end{tabular}
\caption{Main results across two backbones (Qwen2.5-7B-Instruct and Qwen3-4B) and two RLVR algorithms (GRPO and DAPO). All values are percentages. Ours denotes Guided-RLVR. AVG is the unweighted mean of the ten reported metrics. Within each backbone and RLVR algorithm block, the best value in each column is shown in bold; ties are also shown in bold.}
\label{tab:main table}
\end{table*}

\begin{figure}[t]
    \centering
    \includegraphics[width=0.95\linewidth]{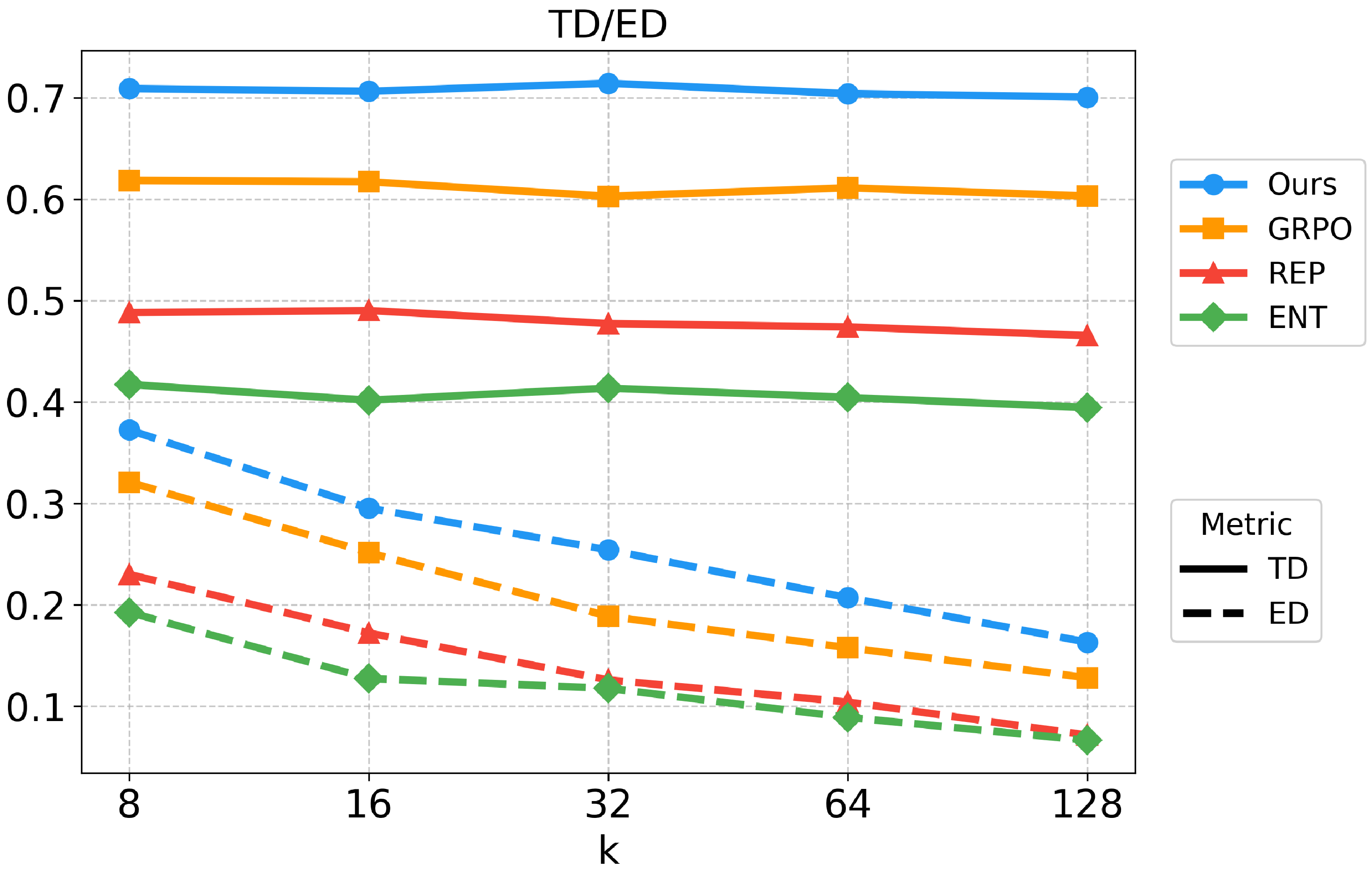}
    \caption{Trajectory diversity across sampling budgets on MATH using Qwen2.5-7B-Instruct with GRPO. Solid lines denote TD and dashed lines denote ED; higher values indicate greater diversity.}
    \label{fig:td_ed}
\end{figure}
\begin{figure*}[t]
    \centering
    \includegraphics[width=0.32\textwidth]{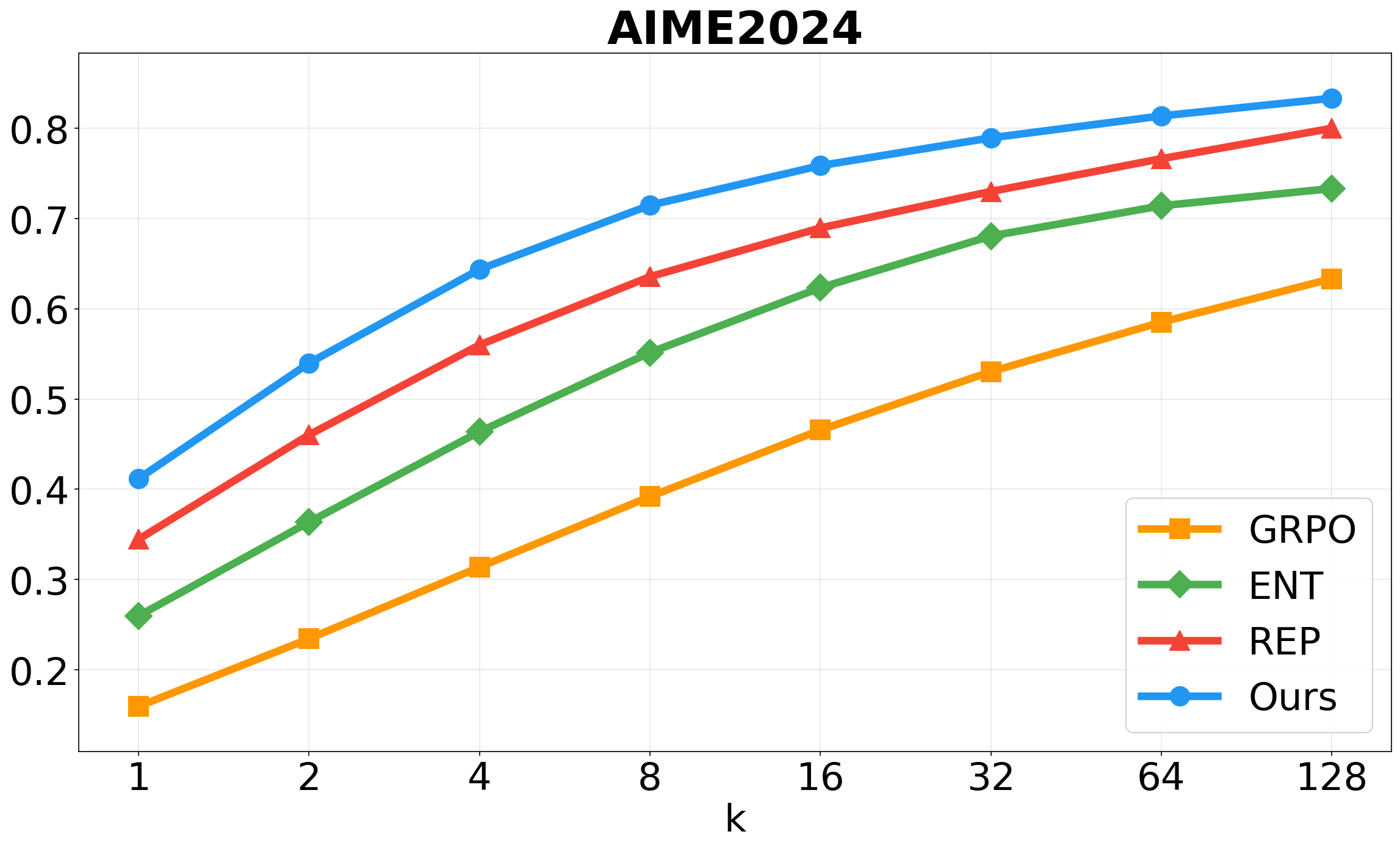}\hfill
    \includegraphics[width=0.32\textwidth]{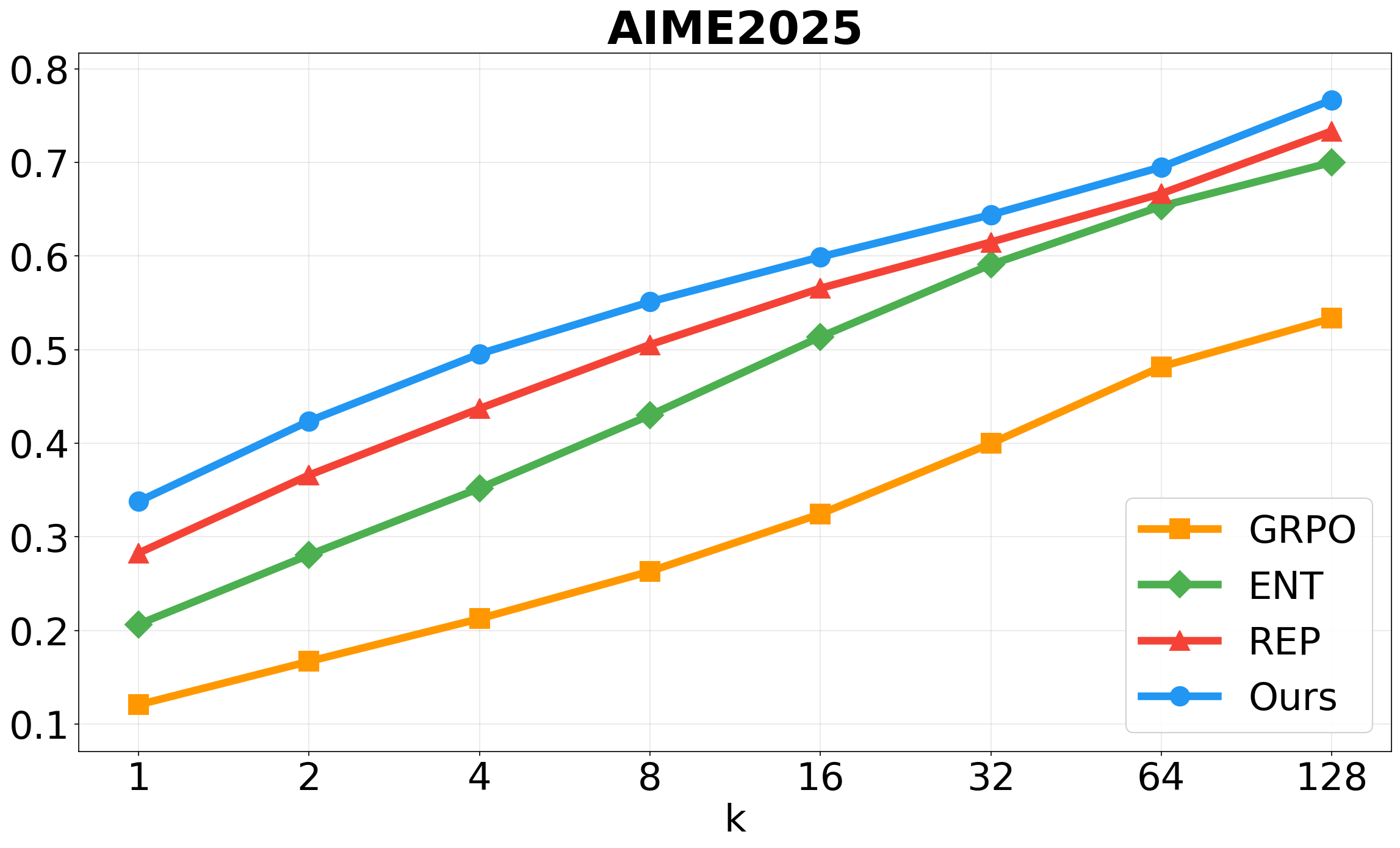}\hfill
    \includegraphics[width=0.32\textwidth]{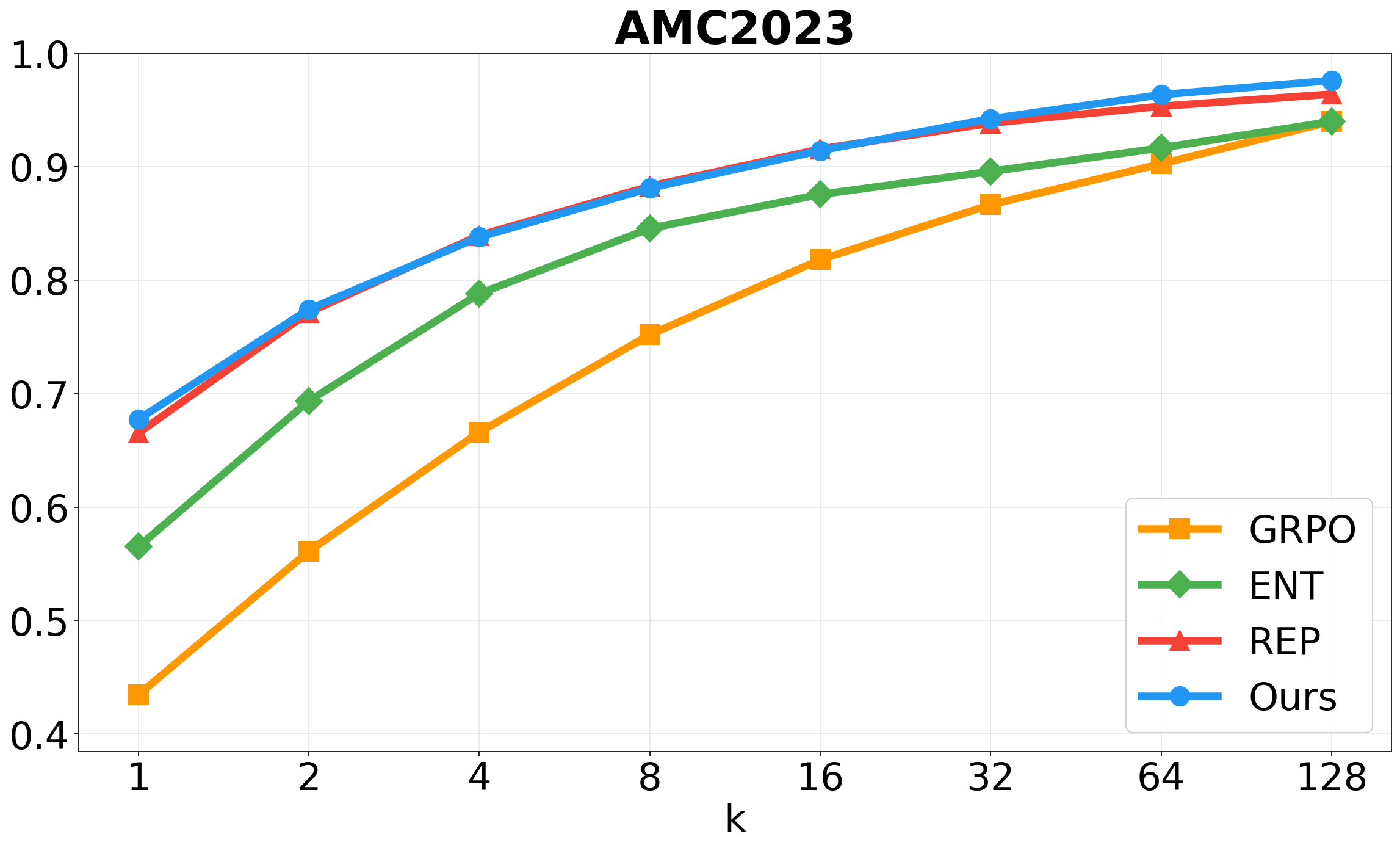}

    \medskip

    \makebox[\textwidth][c]{%
      \includegraphics[width=0.32\textwidth]{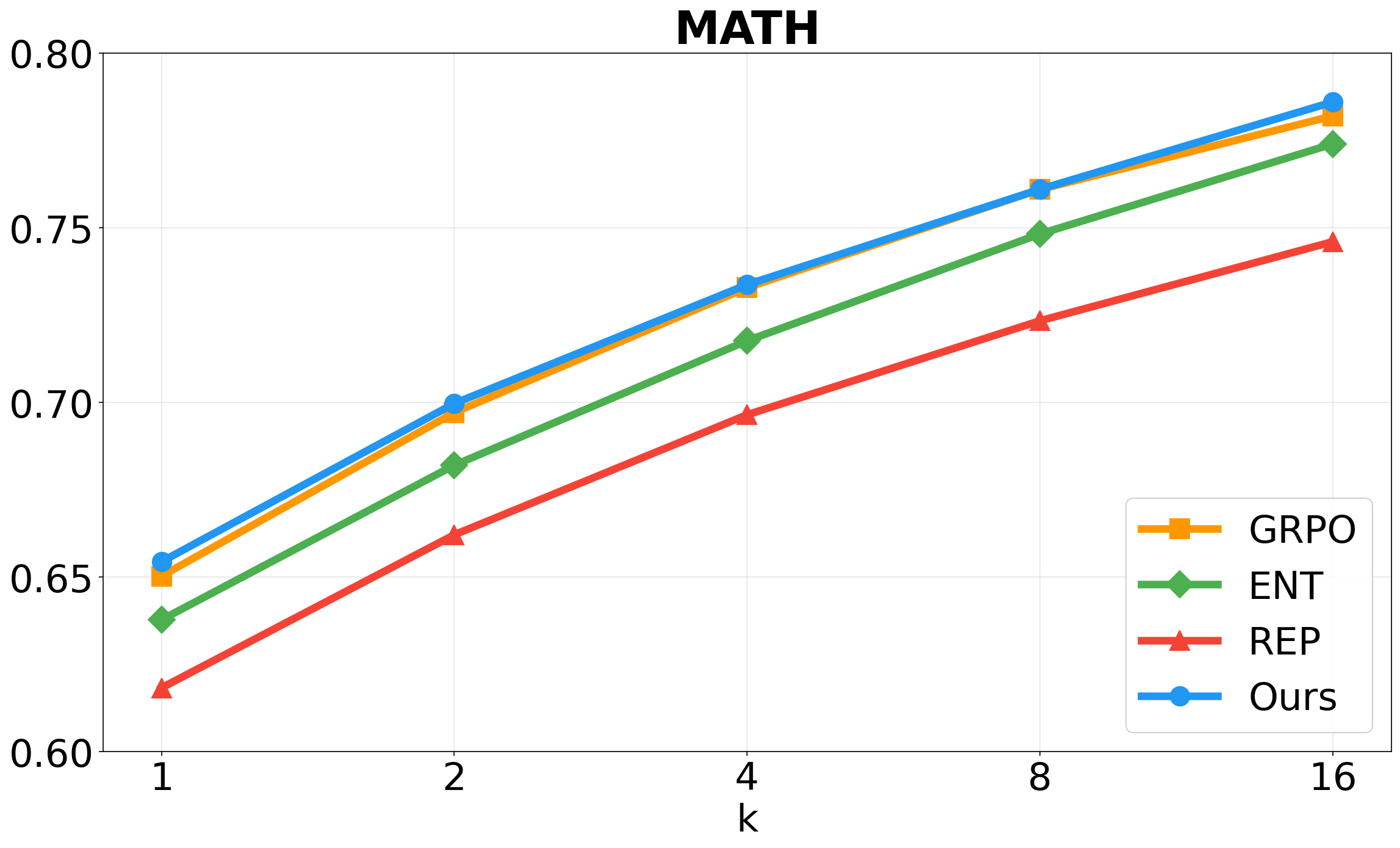}\hspace{0.02\textwidth}%
      \includegraphics[width=0.32\textwidth]{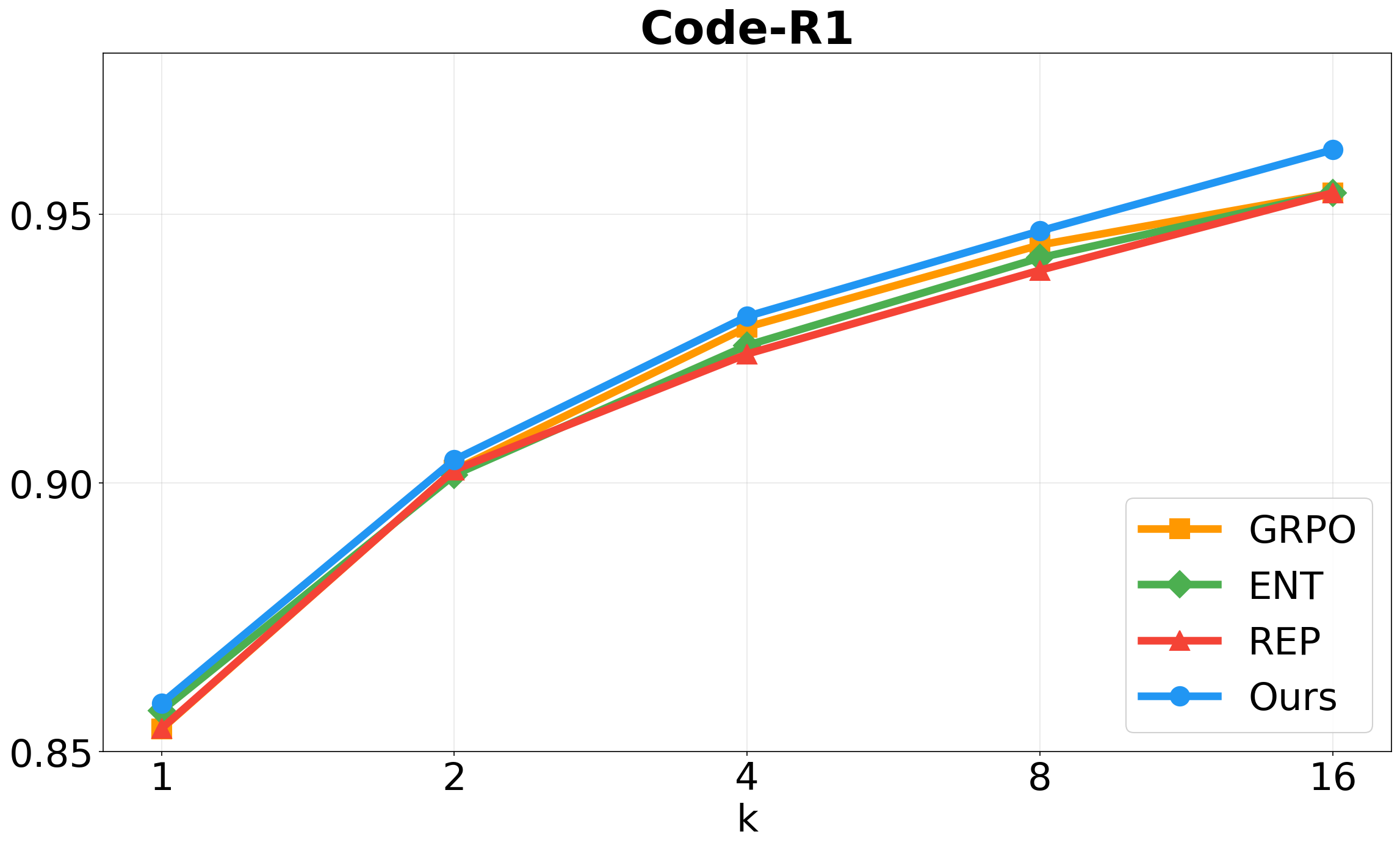}%
    }
    \caption{Pass@$k$ scaling across all five benchmarks using Qwen3-4B with GRPO. The maximum sampling budget is $k{=}128$ for AIME2024, AIME2025, and AMC2023, and $k{=}16$ for MATH and Code-R1.}
    \label{fig:passk}
\end{figure*}

Table 1 presents results across two backbones and five benchmarks. We compare Guided-RLVR with vanilla GRPO and DAPO, as well as variants augmented with ENT or REP. Guided-RLVR achieves the highest average score for every pairing of backbone and RLVR algorithm. Its margin over the strongest baseline ranges from $1.1$ to $2.8$ points.

The benefits of \MethodName{} are most evident under repeated sampling. On AIME2024, AIME2025, and AMC2023, it achieves the best or tied-best pass@128 for every pairing of backbone and RLVR algorithm. With the same budget of 128 samples per problem, a higher pass@128 means that a larger fraction of problems are solved at least once. These results therefore indicate that guided exploration broadens problem coverage.

\subsection{Exploration Analysis}
\label{sec:exploration}

To determine whether the performance gains are accompanied by broader exploration, we analyze trajectory diversity, pass@$k$ scaling, and generation length.
\paragraph{Trajectory diversity.}
We measure diversity on the 500-problem MATH test split using Qwen2.5-7B-Instruct with GRPO. Following~\cite{hu2026diversityincentivized}, we report Textual Diversity (TD) and Equational Diversity (ED). TD is computed as the average pairwise dissimilarity between responses at the sentence level. ED is defined as the proportion of equations in a given response that do not appear in any other response sampled for the same problem. Higher values of TD and ED indicate greater textual variation and greater derivational diversity, respectively.

Figure~\ref{fig:td_ed} reports both metrics for sampling budgets from $k{=}8$ to $k{=}128$. \MethodName{} achieves the highest TD and ED at every budget. At $k{=}128$, it obtains TD$=0.70$ and ED$=0.16$, compared with TD$=0.60$ and ED$=0.13$ for vanilla GRPO. ENT and REP remain below vanilla GRPO on both metrics across all budgets, showing that exploration-oriented objectives do not necessarily yield diverse rollouts. By directly perturbing the sampling distribution through state-conditioned steering token selection, \MethodName{} produces greater textual variation and a broader range of equation sets under the same sampling budget.

\paragraph{Pass@$k$ scaling.}
Table~\ref{tab:main table} reports performance at selected sampling budgets. To examine performance across the full range of budgets, Figure~\ref{fig:passk} presents pass@$k$ curves for all five benchmarks using Qwen3-4B with GRPO. \MethodName{} maintains a clear advantage throughout the sampling range on AIME2024 and AIME2025, and remains best or competitive across budgets on AMC2023, MATH, and Code-R1. The consistent ranking across values of $k$ shows that the gains are not specific to the endpoint budgets reported in Table~\ref{tab:main table}. At any fixed $k$, a higher pass@$k$ means that more problems are solved by at least one sampled response; the curves therefore demonstrate more effective use of the available sampling budget.
\paragraph{Generation length.}

\begin{table}[t]
  \centering
  \small
  \setlength{\tabcolsep}{3pt}
  \renewcommand{\arraystretch}{1.05}
  \begin{tabular}{lcccc}
    \toprule
    \textbf{Method} & \textbf{MATH} & \textbf{AIME2024} & \textbf{AIME2025} & \textbf{AMC2023} \\
    \midrule
    GRPO & \textbf{326.36} & 1224.94 & 935.95 & 818.50 \\
    GRPO + \textbf{ENT} & 413.90 & 1224.84 & 910.15 & 724.80 \\
    GRPO + \textbf{REP} & 407.15 & 1207.24 & 964.91 & 794.88 \\
    \rowcolor{rl-light} GRPO + \textbf{Ours} & 338.04 & \textbf{1087.85} & \textbf{748.06} & \textbf{693.89} \\
    \bottomrule
  \end{tabular}
  \caption{Average generation length in tokens for the GRPO variants on four mathematical benchmarks using Qwen2.5-7B-Instruct, averaged over all completions. Lower values indicate shorter generations.}
  \label{tab:token_consumption}
\end{table}

To examine whether broader exploration simply reflects longer generations, we compare the average generation length of the GRPO variants on four mathematical benchmarks using Qwen2.5-7B-Instruct (Table~\ref{tab:token_consumption}). \MethodName{} remains close to vanilla GRPO on MATH and produces shorter responses on all three competition benchmarks. It also uses fewer tokens than ENT and REP on every benchmark. These results indicate that the improvements in trajectory diversity and pass@$k$ are not primarily explained by longer reasoning chains~\cite{wu2025invisible}. State-conditioned steering redirects the rollout distribution without mechanically extending the generated sequences.

Overall, \MethodName{} enhances exploration in two complementary respects: higher pass@$k$ reflects broader problem coverage under a fixed sampling budget, while higher TD and ED indicate greater textual and equational diversity among sampled trajectories. These benefits are obtained without longer generations, indicating that increased token consumption is not their primary explanation. Appendix~\ref{sec:controller_analysis} further examines the controller's action distribution and token-specific outcomes.

\subsection{Ablation and Sensitivity}
\label{sec:ablation}

\paragraph{Setup.}
We conduct a controlled ablation on the 500-problem MATH test split using Qwen2.5-7B-Instruct trained with GRPO. Starting from the complete \MethodName{} configuration, we modify one component at a time: (a) replace the learned, state-conditioned controller with random steering token selection; (b) always select the fixed steering token ``Wait''; (c) remove the entropy threshold while retaining the sentence-boundary trigger; (d) remove the log-perplexity exploration reward; (e) grant the exploration bonus to all rollouts, in contrast to restricting it to those that are both successfully perturbed and correct; (f) omit supervised controller cold-start training while retaining online joint optimization; or (g) exclude steering-attempt positions from the policy loss while retaining guided generation and controller updates. All variants share the RL training configuration and are evaluated at their respective best checkpoints. We report pass@1 and pass@16 using 16 independent samples per problem. Pass@1 estimates the success probability of a single sampled response, whereas pass@16 reflects the ability to produce at least one correct solution across multiple attempts.

\begin{table}[htbp]
    \centering
    \small
    \setlength{\tabcolsep}{3pt}
    \begin{tabular}{lcc}
        \toprule
        \textbf{Variant} & \textbf{pass@1} & \textbf{pass@16} \\
        \midrule
        Vanilla GRPO & 62.1 & 79.4 \\
        \midrule
        \textbf{\MethodName{}} & \textbf{65.6} & \textbf{80.6} \\
        (a) Random steering token selection & 63.4 & 77.2 \\
        (b) Fixed steering token (``Wait'') & 63.9 & 78.2 \\
        (c) No entropy threshold & 62.1 & 79.2 \\
        (d) No log-perplexity reward & 62.0 & 79.6 \\
        (e) Unconditional bonus & 64.6 & 79.4 \\
        (f) No controller cold start & 64.0 & 79.6 \\
        (g) Mask steering-attempt positions & 63.8 & 80.0 \\
        \bottomrule
    \end{tabular}
    \caption{Component ablation on MATH using Qwen2.5-7B-Instruct with GRPO. All values are percentages. Both pass@1 and pass@16 are estimated from 16 independent samples per problem at the selected checkpoint.}
    \label{tab:ablation_study_advanced}
\end{table}

\paragraph{Steering-token selection.}
Replacing the learned controller with random steering-token selection lowers pass@1 from $65.6$ to $63.4$ and pass@16 from $80.6$ to $77.2$. The decline in both metrics suggests that state-conditioned token selection improves individual rollouts and helps generate useful complementary reasoning trajectories under repeated sampling.

\paragraph{Fixed token selection.}
Always selecting ``Wait'' lowers pass@1 from $65.6$ to $63.9$ and pass@16 from $80.6$ to $78.2$. This gap suggests that a single connective is insufficient to cover the different reasoning transitions needed across problems and rollouts. ``Wait'' promotes reflection but cannot provide other semantic directions such as progression, causality, contrast, or hypothesis. State-conditioned selection among multiple categories therefore provides broader steering and improves problem coverage under repeated sampling.

\paragraph{Entropy threshold.}
Removing the entropy threshold decreases pass@1 by $3.5$ points and pass@16 by $1.4$ points. Restricting interventions to high-entropy sentence boundaries therefore benefits both metrics. During training without the threshold, policy entropy rises instead of converging while evaluation accuracy declines (Appendix~\ref{sec:appendix_entropy_dynamics}). This pattern supports the role of the entropy threshold in controlling perturbations and maintaining stable guided exploration.

\paragraph{Log-perplexity reward.}
Removing the log-perplexity exploration reward decreases pass@1 from $65.6$ to $62.0$ and pass@16 from $80.6$ to $79.6$. Without this auxiliary signal, the model receives no direct feedback on whether a guided reasoning chain makes the correct answer more predictable. The decline in both metrics suggests that the log-perplexity reward helps reinforce useful perturbations beyond the sparse correctness reward alone.

\paragraph{Conditional exploration bonus.}
Granting the exploration bonus unconditionally decreases pass@1 from $65.6$ to $64.6$ and pass@16 from $80.6$ to $79.4$. The log-perplexity bonus measures whether the reasoning chain makes the generated answer more predictable, not whether that answer is correct. An incorrect answer can therefore receive a positive bonus when it is granted unconditionally. Restricting the bonus to successfully perturbed, correct rollouts keeps this auxiliary signal aligned with task correctness.

\paragraph{Controller cold start.}
Without supervised controller initialization, \MethodName{} achieves pass@1 of $64.0$ and pass@16 of $79.60$. These scores exceed vanilla GRPO by $1.94$ and $0.20$ percentage points, respectively. Its pass@16 also exceeds ENT ($78.8$) and REP ($78.0$), although its pass@1 remains below both exploration baselines. Cold start adds $1.56$ points in pass@1 and $1.00$ point in pass@16 over this variant.

\paragraph{Intervention-position loss masking.}
To test whether intervention positions should contribute to policy learning, variant (g) retains guided generation and controller updates but excludes the policy-loss term at every position where steering is attempted, even if the promoted token is not sampled. Generated tokens remain in the context, and GRPO averages over the remaining valid positions. This masking lowers pass@1/pass@16 from $65.6/80.6$ to $63.8/80.0$, favoring retention of intervention positions in the policy loss (Appendix~\ref{sec:appendix_token_mask}).

\subsubsection{Hyperparameter Sensitivity}

\begin{figure}[htbp]
    \centering
    \includegraphics[width=0.95\linewidth]{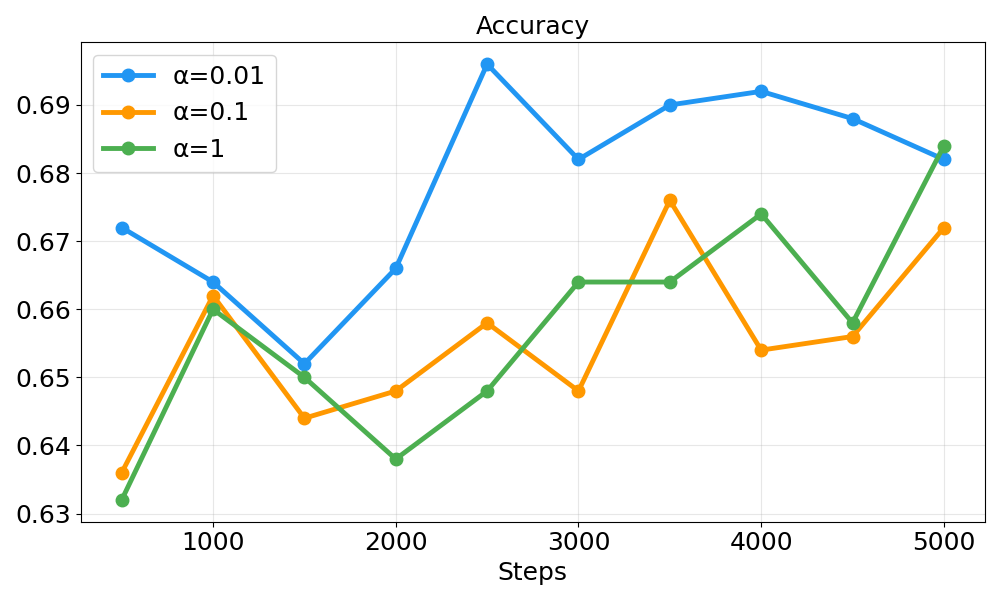}
    \caption{Effect of the log-perplexity reward coefficient $\alpha$ on MATH evaluation accuracy during training using Qwen2.5-7B-Instruct.}
    \label{fig:logperp_ablation}
\end{figure}

We investigate the impact of the log-perplexity exploration bonus coefficient $\alpha$. Figure~\ref{fig:logperp_ablation} compares $\alpha \in \{0.01, 0.1, 1.0\}$ on MATH using Qwen2.5-7B-Instruct. The $\alpha=0.01$ setting achieves the highest peak accuracy and outperforms the larger coefficients at most checkpoints, whereas $\alpha=0.1$ and $\alpha=1.0$ generally produce lower and less stable accuracy. Because the bonus rewards reductions in answer uncertainty, excessive weighting can favor perturbations that make the answer more predictable without improving its correctness. The bonus should therefore remain auxiliary to the verifiable reward, keeping guided exploration aligned with task correctness; we use $\alpha=0.01$ throughout.

The ablations establish that effective guided exploration depends on more than injecting steering tokens. The controller must adapt the semantic direction to the current reasoning state, while entropy-based triggering confines interventions to points where redirection is most useful. The log-perplexity bonus complements sparse task feedback, but must remain properly weighted and conditioned on correctness to avoid rewarding unhelpful reasoning. These components play complementary roles in making exploration both diverse and task-aligned.

\subsection{Computational Cost}
\label{sec:compute_cost}

We measure RL-stage compute cost in a separate timing benchmark using Qwen2.5-7B-Instruct with LoRA on MATH. Each run uses one RTX PRO 6000 Blackwell GPU, $G=10$ generations per prompt, a maximum of 1,024 new tokens, and 5,000 training steps, with evaluation every 500 steps. The recorded \MethodName{} run uses the controller without supervised cold start.

\begin{table}[htbp]
    \centering
    \small
    \setlength{\tabcolsep}{3pt}
    \begin{tabular}{@{}lrr@{}}
        \toprule
        \textbf{Method} & \textbf{GPU-h} & \textbf{s/step} \\
        \midrule
        GRPO & 15.92 & 11.46 \\
        GRPO + ENT & 14.22 & 10.24 \\
        GRPO + REP & 16.55 & 11.92 \\
        \rowcolor{rl-light} GRPO + Ours & 17.92 & 12.90 \\
        \bottomrule
    \end{tabular}
    \caption{RL-stage compute cost under the shared timing configuration. Ours denotes \MethodName{} without supervised controller cold start.}
    \label{tab:compute_cost}
\end{table}

As shown in Table~\ref{tab:compute_cost}, \MethodName{} requires 17.92 GPU-hours, compared with 15.92 for vanilla GRPO, adding 2.00 GPU-hours ($12.6\%$ overhead).

\section{Conclusion}

Standard RLVR reinforces successful trajectories once sampled, yet rarely reaches valid low-prior reasoning paths under a finite rollout budget. We introduced \MethodName{} to enable guided exploration in RLVR. Its state-conditioned controller selects semantically meaningful steering tokens at uncertain reasoning points. By jointly optimizing the controller and policy, \MethodName{} learns where and how to redirect generation toward underexplored reasoning trajectories.

Across five benchmarks, \MethodName{} achieves the highest average performance for every pairing of backbone and RLVR algorithm, with particularly consistent gains at larger sampling budgets. Its higher trajectory diversity and broader problem coverage, achieved without longer generations, support the effectiveness of state-conditioned steering in broadening the sampled reasoning trajectories. Ablations further validate the complementary roles of adaptive token selection, entropy-based intervention, and task-aligned exploration feedback. These findings establish targeted perturbation as an effective approach to exploration in RLVR and motivate future work on learned intervention timing and richer steering spaces.

\section*{Acknowledgments}

This work was funded by the National Natural Science Foundation of China (NSFC) under Grants No. 62388101, No. 62406013, and No. U25B2070, the Beijing Advanced Innovation Center Funds for Future Blockchain and Privacy Computing, and the Fundamental Research Funds for the Central Universities.
\bibliography{aaai2027}

\begin{thebibliography}{33}
\providecommand{\natexlab}[1]{#1}

\bibitem[{Achiam et~al.(2023)Achiam, Adler, Agarwal, Ahmad, Akkaya, Aleman, Almeida, Altenschmidt, Altman, Anadkat et~al.}]{achiam2023gpt}
Achiam, J.; Adler, S.; Agarwal, S.; Ahmad, L.; Akkaya, I.; Aleman, F.~L.; Almeida, D.; Altenschmidt, J.; Altman, S.; Anadkat, S.; et~al. 2023.
\newblock Gpt-4 technical report.
\newblock \emph{arXiv preprint arXiv:2303.08774}.

\bibitem[{Cai et~al.(2025)Cai, Zhao, Liu, Jiang, Zhang, Wu, Hwang, and Li}]{cai-etal-2025-role}
Cai, C.; Zhao, X.; Liu, H.; Jiang, Z.; Zhang, T.; Wu, Z.; Hwang, J.-N.; and Li, L. 2025.
\newblock The Role of Deductive and Inductive Reasoning in Large Language Models.
\newblock In Che, W.; Nabende, J.; Shutova, E.; and Pilehvar, M.~T., eds., \emph{Proceedings of the 63rd Annual Meeting of the Association for Computational Linguistics (Volume 1: Long Papers)}, 16780--16790. Vienna, Austria: Association for Computational Linguistics.
\newblock ISBN 979-8-89176-251-0.

\bibitem[{Chen et~al.(2021)Chen, Tworek, Jun, Yuan, Pinto, Kaplan, Edwards, Burda, Joseph, Brockman et~al.}]{chen2021evaluating}
Chen, M.; Tworek, J.; Jun, H.; Yuan, Q.; Pinto, H. P. D.~O.; Kaplan, J.; Edwards, H.; Burda, Y.; Joseph, N.; Brockman, G.; et~al. 2021.
\newblock Evaluating large language models trained on code.
\newblock \emph{arXiv preprint arXiv:2107.03374}.

\bibitem[{Cheng et~al.(2026)Cheng, Huang, Zhu, Dai, Zhao, Zhang, and Wei}]{cheng2026reasoning}
Cheng, D.; Huang, S.; Zhu, X.; Dai, B.; Zhao, X.; Zhang, Z.; and Wei, F. 2026.
\newblock Reasoning with exploration: An entropy perspective.
\newblock In \emph{Proceedings of the AAAI Conference on Artificial Intelligence}, volume~40, 30377--30385.

\bibitem[{Dai et~al.(2025)Dai, Song, Liu, Liang, Yu, Mi, Tu, Liu, Zheng, Zhu, and Yu}]{dai2025cdecuriositydrivenexplorationefficient}
Dai, R.; Song, L.; Liu, H.; Liang, Z.; Yu, D.; Mi, H.; Tu, Z.; Liu, R.; Zheng, T.; Zhu, H.; and Yu, D. 2025.
\newblock CDE: Curiosity-Driven Exploration for Efficient Reinforcement Learning in Large Language Models.
\newblock arXiv:2509.09675.

\bibitem[{Guha et~al.(2025)Guha, Marten, Keh, Raoof, Smyrnis, Bansal, Nezhurina, Mercat, Vu, Sprague et~al.}]{guha2025openthoughts}
Guha, E.~K.; Marten, R.; Keh, S.; Raoof, N.; Smyrnis, G.; Bansal, H.; Nezhurina, M.; Mercat, J.; Vu, T.; Sprague, Z.~R.; et~al. 2025.
\newblock Openthoughts: Data recipes for reasoning models.
\newblock In \emph{First Workshop on Foundations of Reasoning in Language Models}.

\bibitem[{Guo et~al.(2025)Guo, Yang, Zhang, Song, Wang, Zhu, Xu, Zhang, Ma, Bi et~al.}]{guo2025deepseek}
Guo, D.; Yang, D.; Zhang, H.; Song, J.; Wang, P.; Zhu, Q.; Xu, R.; Zhang, R.; Ma, S.; Bi, X.; et~al. 2025.
\newblock DeepSeek-R1 incentivizes reasoning in LLMs through reinforcement learning.
\newblock \emph{Nature}, 645(8081): 633--638.

\bibitem[{Guo et~al.(2026)Guo, Deng, Cheng, and Tang}]{guo-etal-2026-g2rpo}
Guo, Y.; Deng, W.; Cheng, Z.; and Tang, X. 2026.
\newblock {G}$^2${RPO}-A: Guided Group Relative Policy Optimization with Adaptive Guidance.
\newblock In Liakata, M.; Moreira, V.~P.; Zhang, J.; and Jurgens, D., eds., \emph{Proceedings of the 64th Annual Meeting of the {A}ssociation for {C}omputational {L}inguistics (Volume 1: Long Papers)}, 4525--4539. San Diego, California, United States: Association for Computational Linguistics.
\newblock ISBN 979-8-89176-390-6.

\bibitem[{Hendrycks et~al.(2021)Hendrycks, Burns, Kadavath, Arora, Basart, Tang, Song, and Steinhardt}]{hendrycks2021measuring}
Hendrycks, D.; Burns, C.; Kadavath, S.; Arora, A.; Basart, S.; Tang, E.; Song, D.; and Steinhardt, J. 2021.
\newblock Measuring mathematical problem solving with the math dataset.
\newblock \emph{arXiv preprint arXiv:2103.03874}.

\bibitem[{Hu et~al.(2026)Hu, Zhang, Li, Yan, Hu, Cui, Qu, Chen, Cheng, and Wang}]{hu2026diversityincentivized}
Hu, Z.; Zhang, S.; Li, Y.; Yan, J.; Hu, X.; Cui, L.; Qu, X.; Chen, C.; Cheng, Y.; and Wang, Z. 2026.
\newblock Diversity-Incentivized Exploration for Versatile Reasoning.
\newblock In \emph{The Fourteenth International Conference on Learning Representations}.

\bibitem[{Huang et~al.(2026{\natexlab{a}})Huang, Xu, Wang, Yi, Gong, Li, Xiong, Li, Jiang, and Zhou}]{huang-etal-2026-low}
Huang, G.; Xu, T.; Wang, M.; Yi, Q.; Gong, X.; Li, S.; Xiong, R.; Li, K.; Jiang, Y.; and Zhou, B. 2026{\natexlab{a}}.
\newblock Low-probability Tokens Sustain Exploration in Reinforcement Learning with Verifiable Reward.
\newblock In Liakata, M.; Moreira, V.~P.; Zhang, J.; and Jurgens, D., eds., \emph{Findings of the {A}ssociation for {C}omputational {L}inguistics: {ACL} 2026}, 24158--24188. San Diego, California, United States: Association for Computational Linguistics.
\newblock ISBN 979-8-89176-395-1.

\bibitem[{Huang et~al.(2026{\natexlab{b}})Huang, Tang, Liu, Bai, Yang, and Wu}]{huang-etal-2026-think}
Huang, H.-Y.; Tang, C.; Liu, W.; Bai, C.; Yang, S.; and Wu, Y. 2026{\natexlab{b}}.
\newblock Think Outside the Policy: In-Context Steered Policy Optimization.
\newblock In Liakata, M.; Moreira, V.~P.; Zhang, J.; and Jurgens, D., eds., \emph{Findings of the {A}ssociation for {C}omputational {L}inguistics: {ACL} 2026}, 2758--2776. San Diego, California, United States: Association for Computational Linguistics.
\newblock ISBN 979-8-89176-395-1.

\bibitem[{Jiang et~al.(2025)Jiang, Li, Chen, Liu, Cheng, and Shao}]{jiang2025rethinkingentropy}
Jiang, Y.; Li, Y.; Chen, G.; Liu, D.; Cheng, Y.; and Shao, J. 2025.
\newblock Rethinking Entropy Regularization in Large Reasoning Models.
\newblock arXiv:2509.25133.

\bibitem[{Kwiatkowski et~al.(2026)Kwiatkowski, Butt, Labiad, Kempe, and Ollivier}]{kwiatkowski2026likelihood}
Kwiatkowski, A.; Butt, N.; Labiad, I.; Kempe, J.; and Ollivier, Y. 2026.
\newblock Likelihood-Based Reward Designs for General LLM Reasoning.
\newblock \emph{arXiv preprint arXiv:2602.03979}.

\bibitem[{Lightman et~al.(2023)Lightman, Kosaraju, Burda, Edwards, Baker, Lee, Leike, Schulman, Sutskever, and Cobbe}]{lightman2023let}
Lightman, H.; Kosaraju, V.; Burda, Y.; Edwards, H.; Baker, B.; Lee, T.; Leike, J.; Schulman, J.; Sutskever, I.; and Cobbe, K. 2023.
\newblock Let's verify step by step.
\newblock In \emph{The Twelfth International Conference on Learning Representations}.

\bibitem[{Liu and Zhang(2025)}]{code-r1}
Liu, J.; and Zhang, L. 2025.
\newblock Code-R1: Reproducing R1 for Code with Reliable Rewards.

\bibitem[{Qian et~al.(2026)Qian, Liu, Wen, Bai, Liu, and Shao}]{qian2026demystifying}
Qian, C.; Liu, D.; Wen, H.; Bai, Z.; Liu, Y.; and Shao, J. 2026.
\newblock Demystifying reasoning dynamics with mutual information: Thinking tokens are information peaks in llm reasoning.
\newblock \emph{Advances in Neural Information Processing Systems}, 38: 12533--12572.

\bibitem[{Qwen et~al.(2025)Qwen, :, Yang, Yang, Zhang, Hui, Zheng, Yu, Li, Liu, Huang, Wei, Lin, Yang, Tu, Zhang, Yang, Yang, Zhou, Lin, Dang, Lu, Bao, Yang, Yu, Li, Xue, Zhang, Zhu, Men, Lin, Li, Tang, Xia, Ren, Ren, Fan, Su, Zhang, Wan, Liu, Cui, Zhang, and Qiu}]{qwen2025qwen25technicalreport}
Qwen; :; Yang, A.; Yang, B.; Zhang, B.; Hui, B.; Zheng, B.; Yu, B.; Li, C.; Liu, D.; Huang, F.; Wei, H.; Lin, H.; Yang, J.; Tu, J.; Zhang, J.; Yang, J.; Yang, J.; Zhou, J.; Lin, J.; Dang, K.; Lu, K.; Bao, K.; Yang, K.; Yu, L.; Li, M.; Xue, M.; Zhang, P.; Zhu, Q.; Men, R.; Lin, R.; Li, T.; Tang, T.; Xia, T.; Ren, X.; Ren, X.; Fan, Y.; Su, Y.; Zhang, Y.; Wan, Y.; Liu, Y.; Cui, Z.; Zhang, Z.; and Qiu, Z. 2025.
\newblock Qwen2.5 Technical Report.
\newblock arXiv:2412.15115.

\bibitem[{Shao et~al.(2024)Shao, Wang, Zhu, Xu, Song, Bi, Zhang, Zhang, Li, Wu et~al.}]{shao2024deepseekmath}
Shao, Z.; Wang, P.; Zhu, Q.; Xu, R.; Song, J.; Bi, X.; Zhang, H.; Zhang, M.; Li, Y.; Wu, Y.; et~al. 2024.
\newblock Deepseekmath: Pushing the limits of mathematical reasoning in open language models.
\newblock \emph{arXiv preprint arXiv:2402.03300}.

\bibitem[{Sloman and Sloman(2009)}]{sloman2009causal}
Sloman, S.; and Sloman, S.~A. 2009.
\newblock \emph{Causal models: How people think about the world and its alternatives}.
\newblock Oxford University Press.

\bibitem[{Team et~al.(2024)Team, Georgiev, Lei, Burnell, Bai, Gulati, Tanzer, Vincent, Pan, Wang et~al.}]{team2024gemini}
Team, G.; Georgiev, P.; Lei, V.~I.; Burnell, R.; Bai, L.; Gulati, A.; Tanzer, G.; Vincent, D.; Pan, Z.; Wang, S.; et~al. 2024.
\newblock Gemini 1.5: Unlocking multimodal understanding across millions of tokens of context.
\newblock \emph{arXiv preprint arXiv:2403.05530}.

\bibitem[{Tomihari and Sato(2026)}]{tomihari2026power}
Tomihari, A.; and Sato, I. 2026.
\newblock Power Distribution Bridges Sampling, Self-Reward RL, and Self-Distillation.
\newblock \emph{arXiv preprint arXiv:2605.04542}.

\bibitem[{Tuyls et~al.(2026)Tuyls, Foster, Krishnamurthy, and Ash}]{tuyls2026representationbased}
Tuyls, J.; Foster, D.~J.; Krishnamurthy, A.; and Ash, J.~T. 2026.
\newblock Representation-Based Exploration for Language Models: From Test-Time to Post-Training.
\newblock In \emph{The Fourteenth International Conference on Learning Representations}.

\bibitem[{van~der Maaten and Hinton(2008)}]{JMLR:v9:vandermaaten08a}
van~der Maaten, L.; and Hinton, G. 2008.
\newblock Visualizing Data using t-SNE.
\newblock \emph{Journal of Machine Learning Research}, 9(86): 2579--2605.

\bibitem[{Wang et~al.(2025)Wang, Yu, Gao, Zheng, Liu, Lu, Dang, Chen, Yang, Zhang, Liu, Yang, Zhao, Yue, Song, Yu, Huang, and Lin}]{NEURIPS2025_a797c2d2}
Wang, S.; Yu, L.; Gao, C.; Zheng, C.; Liu, S.; Lu, R.; Dang, K.; Chen, X.-H.; Yang, J.; Zhang, Z.; Liu, Y.; Yang, A.; Zhao, A.; Yue, Y.; Song, S.; Yu, B.; Huang, G.; and Lin, J. 2025.
\newblock Beyond the 80/20 Rule: High-Entropy Minority Tokens Drive Effective Reinforcement Learning for LLM Reasoning.
\newblock In Belgrave, D.; Zhang, C.; Lin, H.; Pascanu, R.; Koniusz, P.; Ghassemi, M.; and Chen, N., eds., \emph{Advances in Neural Information Processing Systems}, volume 38, Main Conference, 115452--115486. Curran Associates, Inc.

\bibitem[{Wei et~al.(2023)Wei, Wang, Liu, Ding, and Zhang}]{wei2023magicoder}
Wei, Y.; Wang, Z.; Liu, J.; Ding, Y.; and Zhang, L. 2023.
\newblock Magicoder: Source code is all you need.
\newblock \emph{arXiv preprint arXiv:2312.02120}, 3.

\bibitem[{Wu et~al.(2025)Wu, Xuan, Lu, Liu, Dong, Harchaoui, and Choi}]{wu2025invisible}
Wu, F.; Xuan, W.; Lu, X.; Liu, M.; Dong, Y.; Harchaoui, Z.; and Choi, Y. 2025.
\newblock The invisible leash: Why rlvr may or may not escape its origin.
\newblock \emph{arXiv preprint arXiv:2507.14843}.

\bibitem[{Yang et~al.(2025)Yang, Li, Yang, Zhang, Hui, Zheng, Yu, Gao, Huang, Lv et~al.}]{yang2025qwen3}
Yang, A.; Li, A.; Yang, B.; Zhang, B.; Hui, B.; Zheng, B.; Yu, B.; Gao, C.; Huang, C.; Lv, C.; et~al. 2025.
\newblock Qwen3 technical report.
\newblock \emph{arXiv preprint arXiv:2505.09388}.

\bibitem[{Yao et~al.(2023)Yao, Yu, Zhao, Shafran, Griffiths, Cao, and Narasimhan}]{yao2023tree}
Yao, S.; Yu, D.; Zhao, J.; Shafran, I.; Griffiths, T.; Cao, Y.; and Narasimhan, K. 2023.
\newblock Tree of thoughts: Deliberate problem solving with large language models.
\newblock \emph{Advances in neural information processing systems}, 36: 11809--11822.

\bibitem[{Yu et~al.(2026)Yu, Zhang, Zhu, Yuan, Zuo, Yue, Dai, Fan, Liu, Liu et~al.}]{yu2026dapo}
Yu, Q.; Zhang, Z.; Zhu, R.; Yuan, Y.; Zuo, X.; Yue, Y.; Dai, W.; Fan, T.; Liu, G.; Liu, L.; et~al. 2026.
\newblock Dapo: An open-source llm reinforcement learning system at scale.
\newblock \emph{Advances in Neural Information Processing Systems}, 38: 113222--113244.

\bibitem[{Yue et~al.(2025)Yue, Chen, Lu, Zhao, Wang, Song, and Huang}]{yue2025does}
Yue, Y.; Chen, Z.; Lu, R.; Zhao, A.; Wang, Z.; Song, S.; and Huang, G. 2025.
\newblock Does reinforcement learning really incentivize reasoning capacity in llms beyond the base model?
\newblock \emph{arXiv preprint arXiv:2504.13837}.

\bibitem[{Zeng et~al.(2026)Zeng, Lu, Li, Zhang, Li, and Ren}]{zeng2026large}
Zeng, Y.; Lu, A.; Li, L.; Zhang, Z.; Li, Y.; and Ren, K. 2026.
\newblock Large Language Models Explore by Latent Distilling.
\newblock \emph{arXiv preprint arXiv:2604.24927}.

\bibitem[{Zhu et~al.(2025)Zhu, Zhao, Yan, He, Chen, and Gui}]{zhu2025soft}
Zhu, Q.; Zhao, R.; Yan, H.; He, Y.; Chen, Y.; and Gui, L. 2025.
\newblock Soft Reasoning: Navigating Solution Spaces in Large Language Models through Controlled Embedding Exploration.
\newblock \emph{arXiv preprint arXiv:2505.24688}.

\end{thebibliography}

\appendix
\section{Closed-Form Derivation of Perturbation Injection}
\label{sec:appendix_perturbation}
\paragraph{Problem Setup}
Assume that at generation step $t$, the base language model produces raw logits
\[
z = (z_1, z_2, \dots, z_{|\mathcal{V}|}) \in \mathbb{R}^{|\mathcal{V}|},
\]
where $|\mathcal{V}|$ is the vocabulary size. The corresponding probability distribution is
\[
p_i = \frac{e^{z_i}}{\sum_{j=1}^{|\mathcal{V}|} e^{z_j}} = \frac{e^{z_i}}{S}, \qquad S = \sum_{j=1}^{|\mathcal{V}|} e^{z_j}.
\]
Now suppose we wish to raise the probability of a specific target token $v_k$ (whose vocabulary index is $k$) to a user-specified target value $p_{\mathrm{target}} \in (0,1)$. We do so by adding a nonnegative bias $\delta \geq 0$ to the corresponding logit:
\[
z_k' = z_k + \delta, \qquad z_i' = z_i \quad (\forall i \neq k).
\]
The perturbed probability distribution is then given by
\[
p_i' = \frac{e^{z_i'}}{S'}, \qquad S' = \sum_{j=1}^{|\mathcal{V}|} e^{z_j'},
\]
and our goal is to enforce
\[
p_k' = p_{\mathrm{target}}.
\]
\paragraph{Derivation of the Closed-Form Solution}
We first express the perturbed normalization constant $S'$ in terms of the original quantities. Observe that
\begin{align*}
S' &= \sum_{j \neq k} e^{z_j} + e^{z_k + \delta} \\
&= (S - e^{z_k}) + e^{z_k} e^{\delta} \\
&= S + e^{z_k}(e^{\delta} - 1).
\end{align*}
The target constraint becomes
\[
\frac{e^{z_k} e^{\delta}}{S + e^{z_k}(e^{\delta} - 1)} = p_{\mathrm{target}}.
\]
Let $x = e^{\delta} \geq 1$. Then
\[
\frac{x e^{z_k}}{S - e^{z_k} + x e^{z_k}} = p_{\mathrm{target}}.
\]
Multiplying both sides by the denominator yields
\[
x e^{z_k} = p_{\mathrm{target}} \bigl(S - e^{z_k} + x e^{z_k}\bigr).
\]
Expanding and collecting all terms containing $x$ on the left-hand side gives
\[
x e^{z_k} - p_{\mathrm{target}} x e^{z_k} = p_{\mathrm{target}}(S - e^{z_k}).
\]
Hence,
\[
x e^{z_k}(1 - p_{\mathrm{target}}) = p_{\mathrm{target}}(S - e^{z_k}),
\]
which implies
\[
x = \frac{p_{\mathrm{target}}}{1 - p_{\mathrm{target}}} \cdot \frac{S - e^{z_k}}{e^{z_k}}.
\]
Therefore,
\[
\delta = \log x = \log\!\left(\frac{p_{\mathrm{target}}}{1 - p_{\mathrm{target}}}\right) + \log\!\left(\frac{S}{e^{z_k}} - 1\right).
\]

\paragraph{Simplified Form}
Note that
\[
\frac{S}{e^{z_k}} = e^{\log S - z_k} = e^{\operatorname{logsumexp}(z) - z_k},
\]
where
\[
\operatorname{logsumexp}(z) = \log\!\left(\sum_{j=1}^{|\mathcal{V}|} e^{z_j}\right).
\]
Substituting this into the previous expression, we obtain the exact solution
\begin{equation}
\delta = \log\!\left(\frac{p_{\mathrm{target}}}{1 - p_{\mathrm{target}}}\right) + \log\!\left(e^{\operatorname{logsumexp}(z) - z_k} - 1\right).
\label{eq:appendix_exact_delta}
\end{equation}

\paragraph{Approximate Form}
In practice, when the target token initially has small probability, we typically have $S \gg e^{z_k}$, equivalently When $e^{\operatorname{logsumexp}(z) - z_k} \gg 1$,
\[
\log\!\bigl(e^{\operatorname{logsumexp}(z) - z_k} - 1\bigr) \approx \operatorname{logsumexp}(z) - z_k .
\]
This yields the approximation
\begin{equation}
\delta \approx \log\!\left(\frac{p_{\mathrm{target}}}{1 - p_{\mathrm{target}}}\right) + \operatorname{logsumexp}(z) - z_k.
\label{eq:appendix_approx_delta}
\end{equation}

\paragraph{Nonnegative Clipping}
To ensure that we only increase the target token's probability and never reduce it, we apply the clipping rule
\[
\delta = \max(0, \delta_{\mathrm{approx}}),
\]
where $\delta_{\mathrm{approx}}$ is given by Equation~\eqref{eq:appendix_approx_delta}. When the target token's original probability already exceeds $p_{\mathrm{target}}$, the computed $\delta$ is nonpositive, and clipping sets it to zero, corresponding to no intervention.

\paragraph{Final Perturbation Formula}
Combining the above derivation, the perturbed logits can be written as
\[
\tilde{z}_i =
\begin{cases}
z_i + \delta, & i = k, \\
z_i, & i \neq k,
\end{cases}
\]
with
\[
\delta = \max\!\left(0, \log\!\left(\frac{p_{\mathrm{target}}}{1 - p_{\mathrm{target}}}\right) + \operatorname{logsumexp}(z) - z_k\right).
\]
Writing \(\Delta_t = \operatorname{logsumexp}(z_t) - [z_t]_{i_k}\) for the bias applied at
generation step \(t\) to the target token \(v_k\) (vocabulary index \(i_k\)), the perturbed
logits are \(\tilde{z}_t = z_t + \delta \cdot e_{i_k}\), where \(e_{i_k}\) is the one-hot
vector for \(v_k\). The outer maximum ensures we never decrease the target token's logit,
only boosting it when necessary. Applying the standard decoding pipeline, the perturbed
rollout policy is
\[
\pi^{\mathrm{pert}}_{\theta,\phi}(y_t \mid x, y_{<t}) = \operatorname{softmax}\!\left(\frac{\tilde{z}_t}{T}\right),
\]
followed by top-\(k\) and top-\(p\) truncation.

\section{\ControllerName{} Details}
\label{sec:appendix_controller_details}

\subsection{Steering Token Vocabulary}
\label{sec:appendix_steering_vocabulary}
The controller selects a steering token category \(v_k\) from a predefined set \(\mathcal{T} = \{v_1, \dots, v_K\}\) (\(K=6\)), organized by
semantic function. Each category corresponds to one discrete action \(a_t^c = k\) in the controller's action space \(\mathcal{A}\). When a category maps to multiple surface tokens, the specific token to inject is sampled in proportion to that
token's frequency among the DeepSeek-R1 traces used to cold-start the controller, so that more common connectives
are injected more often while rarer variants remain reachable.

  \begin{table}[h]
  \centering
  \small
  \setlength{\tabcolsep}{4pt}
  \begin{tabular}{@{}>{\raggedright\arraybackslash}p{0.28\columnwidth}>{\raggedright\arraybackslash}p{0.28\columnwidth}>{\raggedright\arraybackslash}p{0.37\columnwidth}@{}}
  \toprule
  \textbf{Category} & \textbf{Tokens} & \textbf{Semantic function} \\
  \midrule
  Reflection        & \textit{Wait}                      & Revisit or reconsider a step \\
  Progression       & \textit{Then}                      & Advance to the next step \\
  Causality         & \textit{Therefore}, \textit{Hence} & Draw a consequence \\
  Causal Reflection & \textit{Since}, \textit{Because}   & Justify by a prior cause \\
  Contrast          & \textit{But}, \textit{However}     & Introduce an opposing turn \\
  Hypothesis        & \textit{If}                        & Pose a conditional assumption \\
  \bottomrule
  \end{tabular}
  \caption{Steering token vocabulary \(\mathcal{T}\), grouped by the six semantic categories the controller predicts. Within a category, the injected
  token is drawn in proportion to its frequency in the cold-start DeepSeek-R1 traces.}
  \label{tab:steering-tokens}
  \end{table}

\subsection{Trigger Conditions and Entropy-Threshold Schedule}
\label{sec:appendix_trigger_schedule}
The controller is queried only when three conditions hold: (i) the previous token is a
punctuation mark, corresponding to a sentence boundary; (ii) at least \(M_{\min}\) tokens
have been generated since the last real injection; and (iii) the entropy of the base
model's next-token distribution exceeds a dynamic, per-sequence threshold
\(\theta_{\mathrm{ent}}\) derived from recent entropy statistics and growing progressively
stricter with training progress. For each sequence we maintain a sliding window
\(\mathcal{H}_i\) of the last \(W\) entropy values observed at punctuation positions
(\(W = 3\) in our experiments). At global training step \(g_s\), the threshold for the
\(i\)-th sequence is
\[
\theta_{\mathrm{ent}}^{(i)} = \max(\mathcal{H}_i) + \sigma \cdot \Bigl(
\operatorname{std}(\mathcal{H}_i) + \frac{g_s}{G_{\max}} \cdot
\overline{\operatorname{std}(\mathcal{H})} \Bigr),
\]
where \(\sigma\) is a fixed scale factor, \(G_{\max}\) is the total number of training
steps, and \(\overline{\operatorname{std}(\mathcal{H})}\) is the batch-wise average of the
window standard deviations. A warm-up phase suppresses interventions until each sequence
has accumulated at least \(W\) punctuation events, so the threshold is well-calibrated
before any perturbation. These conditions confine interventions to natural restart points
under genuine uncertainty, while the progressively increasing threshold reduces
unnecessary perturbations as training stabilizes.

\subsection{Hidden-State Aggregation}
\label{sec:appendix_hidden_states}
Let \(h_t^{(l)}\) be the hidden state of the \(l\)-th layer at generation step \(t\) for
the last generated token position. The controller aggregates the last \(L\) layers by
summation:
\[
s_t^c = \sum_{l = N-L+1}^{N} h_t^{(l)} \in \mathcal{S}^c,
\]
where \(N\) is the total number of layers and \(L\) is a hyperparameter.

\section{Training Details}
\label{sec:appendix_training}

\subsection{Cold-Start Training}
\label{sec:appendix_cold_start}
The controller does not classify every token during cold-start training. We segment each DeepSeek-R1 trace at sentence-final punctuation (\texttt{.\,?\,!}), and at each boundary take the hidden state of the boundary token, $s_c=\sum_{l=N-L+1}^{N} h^{(l)}$, as input to the MLP. The MLP then predicts which of the six steering categories should follow at that break: Reflection, Progression, Causality, Causal Reflection, Contrast, or Hypothesis.

Supervision is obtained by rule-based labeling: the category of the steering token that actually appears after each boundary in the R1 trace provides the target. For example, \textit{Wait} $\rightarrow$ Reflection, \textit{Therefore}/\textit{Hence} $\rightarrow$ Causality. We build a class-balanced training set of 100,000 boundary samples per category (600K in total) and train the MLP with a cross-entropy loss for 10 epochs at a learning rate of $5\times10^{-5}$. The no-cold-start ablation omits this pretraining while retaining the same steering vocabulary and within-category token sampler.

During the main RL training stage, the controller is not frozen but continues to be updated online with the policy via the group-relative advantage and the log-perplexity exploration bonus. Its update retains valid events where the promoted token was actually sampled, averages their contributions within each completion, and then averages across completions with at least one such event. Hidden states and advantages are detached for this update. The classification MLP $\mu_\phi$ is optimized with its own optimizer, decoupled from the policy update.

\subsection{Policy Loss Evaluation}
\label{sec:policy_update_details}
The sampled completion, including any steering tokens, is kept unchanged when computing the policy loss. Both probabilities in Equation~\eqref{eq:guided_base_ratio} are evaluated on the generated prefixes using temperature-scaled LLM logits, without the controller's boost or top-$k$/top-$p$ truncation. The ratio compares the current and old unboosted policies; it does not correct for the guided sampling distribution.

GRPO averages the clipped terms in Equation~\eqref{eq:guided_llm_objectives} over valid tokens within each completion and then over completions. DAPO instead normalizes their sum by the generation batch's completion-token count, recorded before validity masking. No KL penalty is applied.

\subsection{Controller Learning Rates and Intervention Details}
\label{sec:appendix_controller_training}
Because the controller's stability requirements differ across tasks, we use a controller learning rate of 5e-7 on MATH and Code-R1, and 5e-6 on AIME2024/2025 and AMC2023. The entropy coefficient \(\lambda_H\) is held at \(\lambda_H^{\text{start}}\) for the first half of training, then annealed linearly to zero. The controller update is skipped when no valid successful intervention is available.

When the controller intervenes, the selected steering token's logit is boosted so that its probability reaches a fixed target $p_{\text{target}}=0.5$. The closed-form derivation of the logit adjustment $\delta$ is provided in Appendix~\ref{sec:appendix_perturbation}. After applying $\delta$, the perturbed distribution is re-normalized with the training temperature and top-$k$/top-$p$ truncation.

For the exploration baselines, we adopt the entropy coefficient of ENT and the representation-bonus scale of REP as specified by their official implementations.

\subsection{Log-Perplexity Reward Parameters}
\label{sec:appendix_reward_parameters}
The transformation $f(\cdot)$ in the main text is defined as
\[
f(\Delta\ell) = \operatorname{clip}\bigl(\operatorname{sigmoid}(\Delta\ell') - k,\; r_{\min}, r_{\max}\bigr),
\]
where $\Delta\ell'$ is obtained by standardizing $\Delta\ell$ within each generation group.
We set $k=0.5$ so that a neutral perturbation ($\Delta\ell'=0$) receives no extra
reward. The shifted term $\mathrm{sigmoid}(\Delta\ell')-k$ is clipped to
$[r_{\min},r_{\max}]=[0,0.2]$ to prevent extreme values, and the resulting bonus
is scaled by $\alpha=0.01$ before being added to the task reward. This bonus is
applied only to the math tasks; for code generation we optimize the correctness
reward alone.

Both $\ell_{\text{base}}$ and $\ell_{\text{cot}}$ are computed on the \emph{same} answer sequence $y$: $\ell_{\text{cot}}$ is
evaluated in the context actually used during generation (with the controller-guided reasoning chain $\hat{x}_{\text{think}}$ preceding the
answer), whereas $\ell_{\text{base}}$ is a counterfactual re-evaluation of that same answer with $\hat{x}_{\text{think}}$ removed from the
context. This requires a single additional forward pass rather than a separately sampled rollout. The gap $\Delta\ell = \ell_{\text{base}} - \ell_{\text{cot}}$
therefore isolates the contribution of the reasoning chain to the model's certainty over the answer.

\subsection{Data Processing and Reward Design}
\label{sec:appendix_data_rewards}
For each dataset example, we construct an instruction-style prompt that enforces a strict output format. Given a problem $x$, we produce a prompt:
\begin{quote}
  \small
  \texttt{You are an expert in math. Please adhere to the following format:}\\
  \texttt{1. Reason through the problem inside <think> </think> tags.}\\
  \texttt{2. Provide the final numerical answer inside <answer> </answer> tags.}\\
  \texttt{3. All fractions, formulas must be in LaTeX format.}\\
  \texttt{4. End your response with <eos>.}\\
  \texttt{Here is an example: '<think> ... </think><answer> ... </answer><eos>'}\\
  \texttt{Problem: \{problem\}}\\
  \texttt{<think>}
\end{quote}
We store the prompt in a \texttt{prompt} field and the ground-truth answer in an \texttt{answer} field.

During training and evaluation, prompts are tokenized with padding and truncation. We set left-padding to align variable-length prompts in a batch. To compute the correctness reward, we parse the substring inside \texttt{<answer>...</answer>} using a regular expression. We treat answers as correct if they match the target string case-insensitively, and additionally attempt numeric matching by casting to floating point and accepting a small tolerance when applicable.

\paragraph{Reward Design}
We use an automatic verification reward composed of:
(i) a format reward that checks whether the completion matches the required \texttt{<think>} and \texttt{<answer>} structure with an additional penalty if there is extra trailing text after the tags, and
(ii) a correctness reward that checks whether the extracted final answer matches the ground truth.
The final reward is a weighted sum of the two components, and we optionally include a soft overlength punishment that discourages excessively long outputs for DAPO.

\subsection{Training Hyperparameters}
\label{sec:appendix_hyperparameters}
Table~\ref{tab:train_hparams} provides the complete training configuration for each backbone model and dataset combination. These model-specific and dataset-specific hyperparameters were tuned to balance training stability, memory efficiency, and generation quality. All other hyperparameters (sampling temperature, $\beta$, clipping range, number of rollouts, learning rate) are held fixed across all configurations and summarized in Table~\ref{tab:train_hyperparams}.

\begin{table}[h]
\centering
\small
\begin{tabular}{p{0.65\linewidth}p{0.25\linewidth}}
  \toprule
  \textbf{Hyperparameter} & \textbf{Value} \\
  \midrule
  Sampling temperature (training) & 1.0 \\
  Sampling temperature (evaluation) & 0.7 \\
  KL penalty coefficient ($\beta$) & 0 \\
  Decoupled clipping (high) & 0.28 \\
  Decoupled clipping (low) & 0.2 \\
  Rollouts per prompt ($G$) & 10 \\
  Learning rate (policy) & 5e-6 \\
  \bottomrule
\end{tabular}
\caption{Training hyperparameters shared across all methods and datasets.}
\label{tab:train_hyperparams}
\end{table}

\begin{table}[htbp]
    \centering
    \small
    \setlength{\tabcolsep}{2pt}        
    \renewcommand{\arraystretch}{1.0}
    \begin{tabular}{@{}lccc@{}}
        \toprule
        \textbf{Dataset} & \textbf{Token Length} & \textbf{Batch Size} & \textbf{Train Batch} \\
        \midrule
        \multicolumn{4}{@{}l}{\textbf{Qwen2.5-7B-Instruct}} \\
        \midrule
        MATH & 1024 & 40 & 4 \\
        AIME2024/2025 & 8192 & 100 & 20 \\
        AMC2023 & 8192 & 100 & 20 \\
        Code-R1 & 8192 & 40 & 4 \\
        \midrule
        \multicolumn{4}{@{}l}{\textbf{Qwen3-4B}} \\
        \midrule
        MATH & 1024 & 40 & 4 \\
        AIME2024/2025 & 8192 & 100 & 20 \\
        AMC2023 & 8192 & 100 & 20 \\
        Code-R1 & 8192 & 40 & 4 \\
        \bottomrule
    \end{tabular}
    \caption{Detailed training configuration for each backbone model and dataset.}
    \label{tab:train_hparams}
\end{table}

\section{Evaluation Protocol}
\label{sec:appendix_evaluation}

\subsection{Pass@k Estimator}
\label{sec:appendix_passk}
All pass@$k$ results are computed with the unbiased estimator of \citet{chen2021evaluating},
\[
\text{pass@}k = \mathbb{E}_{x}\left[1 - \frac{\binom{n-c(x)}{k}}{\binom{n}{k}}\right],
\]
where $c(x)$ is the number of correct samples among the $n$ generated for problem $x$, with $n=128$ for AIME2024/2025 and AMC2023 and $n=16$ for MATH and Code-R1.

\subsection{Sampling Budget $K$}
\label{sec:appendix_sampling_budget}
Table~\ref{tab:eval_k} summarizes, for each benchmark, the number of test problems, the qualitative difficulty, the sampling budget $K$ used for pass@$K$ evaluation, and the rationale behind that choice.
\begin{table}[htbp]
    \centering
    \small
    \setlength{\tabcolsep}{4pt}
    \renewcommand{\arraystretch}{1.05}
    \begin{tabular}{lclc}
        \toprule
        \textbf{Benchmark} & \textbf{\# Problems} & \textbf{Difficulty} & \textbf{$K$} \\
        \midrule
        AIME2024 / 2025 & 30 & Extremely hard & 128 \\
        AMC2023 & 83 & Hard & 128 \\
        MATH & 500 & Moderate & 16 \\
        Code-R1 & 500 & Moderate & 16 \\
        \bottomrule
    \end{tabular}
    \caption{Sampling budget $K$ used for pass@$K$ evaluation on each benchmark.}
    \label{tab:eval_k}
\end{table}
Benchmarks with few, extremely hard problems (AIME, AMC) require a larger $K$ to reveal how much of the solution space a method can actually reach, whereas benchmarks with many, moderately difficult problems (MATH, Code-R1) are already well-estimated at $K=16$.

\section{Additional Results}
\label{sec:appendix_additional_results}

\subsection{Controller Behavior}
\label{sec:controller_analysis}

\begin{figure*}[t]
    \centering
    \begin{tabular}{@{}c@{\hspace{0.02\textwidth}}c@{}}
        \vtop{\hbox{\begin{minipage}{0.49\textwidth}
        \centering
        \includegraphics[width=\linewidth]{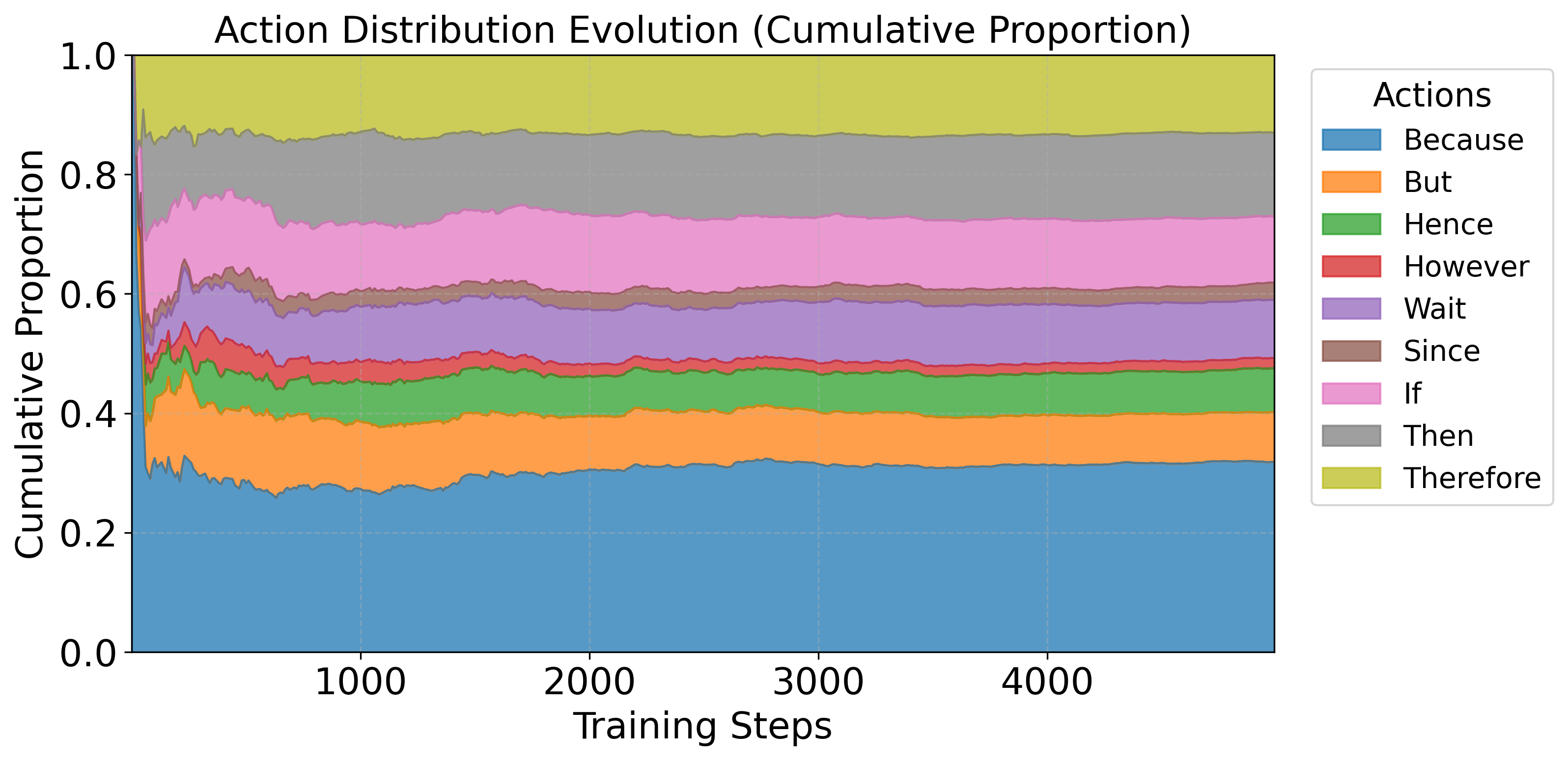}
        \par\smallskip
        \small (a) Action distribution over training.
        \end{minipage}}}
        &
        \vtop{\hbox{\begin{minipage}{0.49\textwidth}
        \centering
        \footnotesize
        \setlength{\tabcolsep}{3pt}
        \begin{tabular}{@{}l@{\hspace{2em}}r@{\hspace{2em}}r@{}}
            \toprule
            \textbf{Token} & \textbf{$n$} & \textbf{Acc.} \\
            \midrule
            Because   & 488 & 70.29 \\
            But       & 127 & 58.27 \\
            Hence     & 116 & 69.83 \\
            However   & 25  & 72.00 \\
            Wait      & 146 & 71.92 \\
            Since     & 42  & 73.81 \\
            If        & 172 & 67.44 \\
            Then      & 209 & 66.03 \\
            Therefore & 198 & 72.22 \\
            \bottomrule
        \end{tabular}
        \par\smallskip
        \small (b) Post-intervention accuracy (\%).
        \end{minipage}}}
    \end{tabular}
    \caption{Controller behavior on MATH using Qwen2.5-7B-Instruct with GRPO. Panel (a) shows the cumulative proportion of each steering-token action over training steps. Panel (b) reports the number of interventions and the subsequent rollout accuracy for each token. The accuracy values are descriptive rather than causal because tokens are selected in different reasoning states and have unequal sample counts.}
    \label{fig:controller_behavior}
\end{figure*}

We examine the behavior of the learned \ControllerName{} in Figure~\ref{fig:controller_behavior}. After an initial adjustment period, the action distribution stabilizes without collapsing to a single token. All nine tokens remain active, with ``Because'' accounting for approximately one third of the selections. Post-intervention accuracy also varies across tokens, indicating that steering actions are associated with different outcomes. Together with the random-selection ablation, these observations show that the gains do not arise from token injection alone and that the controller maintains a non-degenerate action set throughout training.

\subsection{Entropy-Threshold Training Dynamics}
\label{sec:appendix_entropy_dynamics}
We examine how the entropy threshold affects training dynamics on MATH using
Qwen2.5-7B-Instruct with GRPO. We compare the full \MethodName{} configuration with
ablation (c) in Table~\ref{tab:ablation_study_advanced}, which removes the entropy
threshold while retaining the sentence-boundary trigger and the other training
components. In the full configuration, intervention eligibility follows the dynamic
threshold described in Appendix~\ref{sec:appendix_trigger_schedule}.

Figure~\ref{fig:entropy_threshold_dynamics} shows that entropy initially decreases in
both runs, but their subsequent behavior differs substantially. Without the threshold,
entropy rises overall after the early decrease, with a temporary reversal in the middle
of training, and ends well above its initial level. With the threshold, entropy remains
lower and fluctuates within a comparatively narrow range during later training. This
contrast accompanies the performance gap in Table~\ref{tab:ablation_study_advanced}:
removing the threshold lowers pass@1 from $65.6\%$ to $62.1\%$ and pass@16 from
$80.6\%$ to $79.2\%$ at the respective selected checkpoints.

The higher training entropy of the ablated variant therefore does not translate into
better solution coverage under the evaluated sampling budgets. By restricting steering
to sentence boundaries with unusually high next-token uncertainty, the threshold makes
interventions more selective. The observed entropy dynamics and pass@$k$ results are
consistent with this selectivity helping to control perturbations while preserving
useful exploration. The entropy curves alone, however, do not establish how the
diversity or correctness of reasoning trajectories changes during training.

\begin{figure}[tbp]
    \centering
    \includegraphics[width=\linewidth]{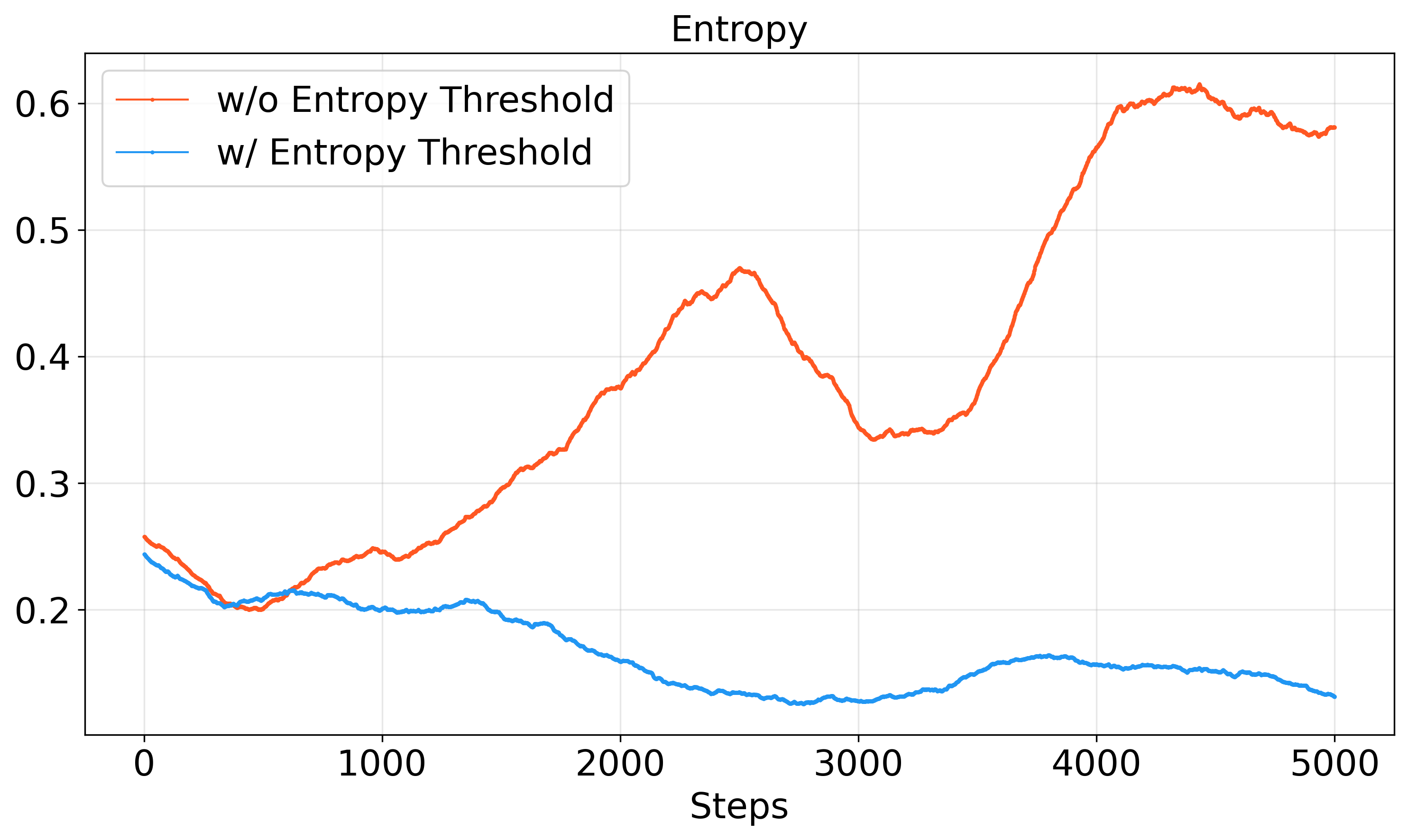}
    \caption{Training entropy on MATH using Qwen2.5-7B-Instruct with GRPO. Removing the entropy threshold (orange) leads to an overall increase after the initial decrease; the full \MethodName{} configuration (blue) maintains lower entropy during later training.}
    \label{fig:entropy_threshold_dynamics}
\end{figure}
\subsection{Intervention-Position Loss Masking}
\label{sec:appendix_token_mask}
The full \MethodName{} retains the policy loss at all valid completion positions (\emph{Full loss}), while the ablation excludes steering-attempt positions (\emph{Masked loss}). Let $d_{i,t}$ record every steering attempt on a non-baseline completion, regardless of whether the promoted token is sampled, and let $v_{i,t}$ be the completion-validity mask. The loss masks are
\begin{equation}
\begin{aligned}
k_{i,t}^{\mathrm{full}}&=v_{i,t},\\
k_{i,t}^{\mathrm{masked}}&=v_{i,t}(1-d_{i,t}).
\end{aligned}
\label{eq:guided_loss_masks}
\end{equation}
This trigger mask differs from the successful-hit filter used in the controller update.

The ablation excludes intervention positions from the policy loss. GRPO averages over the remaining valid tokens, changing both term inclusion and weighting, whereas DAPO retains its generation-token normalizer (Appendix~\ref{sec:policy_update_details}). Both variants use the unboosted policy ratio in Equation~\eqref{eq:guided_base_ratio}; neither substitutes guided sampling probabilities into the denominator.

On MATH using Qwen2.5-7B-Instruct with GRPO, \emph{Masked loss} achieves pass@1 of $63.8$ and pass@16 of $80.0$, compared with $65.6$ and $80.6$ for \emph{Full loss}. The decreases are $1.8$ and $0.6$ percentage points, respectively. Guided generation and the controller update are retained; later token losses still condition on the generated prefixes. The comparison supports the full method's loss inclusion and weighting together, without isolating the contribution of successfully emitted steering tokens alone or correcting the guided behavior-distribution mismatch. It also does not establish the frequency of steering-token use during controller-free inference.

\section{Theoretical Analysis}
\label{sec:appendix_theory}

\subsection{Gradient Bound for Low-Prior Trajectories}
\label{sec:appendix_gradient_bound}
Because the gradient of Equation~\eqref{eq:grpo_obj} is an expectation over the unperturbed rollout distribution \(b_{\mathrm{old}}\), we can split it into contributions
from outside and inside \(\mathcal{A}_x\):
\begin{align*}
\nabla_\theta \mathcal{J}(\theta)
={} & \mathbb{E}_{y \notin \mathcal{A}_x}\bigl[\nabla_\theta \ell(\theta; x, y)\bigr]\,
\bigl(1 - b_{\mathrm{old}}(\mathcal{A}_x \mid x)\bigr) \\
& + \mathbb{E}_{y \in \mathcal{A}_x}\bigl[\nabla_\theta \ell(\theta; x, y)\bigr]\,
b_{\mathrm{old}}(\mathcal{A}_x \mid x),
\end{align*}
where \(\ell(\theta; x, y)\) is the per-trajectory surrogate term in Equation~\eqref{eq:grpo_obj}. Under standard regularity assumptions, rewards,
the clipped importance ratio, and \(\nabla_\theta \log \pi_\theta\) are all bounded, so the per-trajectory gradient norm is bounded by some constant
\(C < \infty\) independent of \(y\); by the triangle inequality, the second term is then bounded by \(C \cdot b_{\mathrm{old}}(\mathcal{A}_x \mid x)\).
Hence whenever \(b_{\mathrm{old}}(\mathcal{A}_x \mid x) \ll 1/G\) or is zero under top-\(p\) truncation, the learning signal from
\(\mathcal{A}_x\) vanishes, mirroring the sampling bound discussed earlier.

\end{document}